\documentclass[11pt,a4paper]{article}
\usepackage{templates/uniai/uniai}

\newif\ifsurveyappendices
\surveyappendicesfalse

\newcommand{\papertitle}{Accelerating Inference in Diffusion and Flow-Matching Models: A Cross-Task Survey from Sampling to System Execution}

\newcommand{\papertype}{}
\newcommand{\paperauthors}{%
  Huanlin~Gao\textsuperscript{1,2},
  Yantao~Li\textsuperscript{1,2,3},
  Shuting~Liu\textsuperscript{1,2},
  Xueqiang~Guo\textsuperscript{1,2},
  Xinpei~Su\textsuperscript{1,2},
  Fang~Zhao\textsuperscript{1,2,\textdagger},
  Kai~Wang\textsuperscript{1,2,\textdaggerdbl},
  Shiguo~Lian\textsuperscript{1,2,\textdagger}%
}
\newcommand{\paperauthornotes}{%
  \textsuperscript{\textdagger}Corresponding author.\enspace
  \textsuperscript{\textdaggerdbl}Project leader.%
}
\newcommand{\paperaffiliations}{%
  \textsuperscript{1}Data Science \& Artificial Intelligence Research Institute, China Unicom\par
  \textsuperscript{2}Unicom Data Intelligence, China Unicom\par
  \textsuperscript{3}National Key Laboratory for Novel Software Technology, Nanjing University%
}
\newcommand{\releaseinfo}{}
\newcommand{\paperlinks}{}
\hypersetup{pdftitle={\papertitle},pdfsubject={Research Survey}}

\begin{document}
\makeunicomtitle
\begin{unicomabstract}
Diffusion and flow-matching models underpin generation across images, video, audio, 3D content, world models, and text, yet their iterative inference incurs coupled bottlenecks across evaluation count, per-step computation, memory traffic, and system execution. Reductions in algorithmic work, such as fewer model evaluations or lower attention density, rarely yield proportional end-to-end speedups. Acceleration methods also often fail to transfer across tasks with distinct structural dependencies. To bridge algorithmic savings and deployment performance, this survey reviews over 300 methods under a mechanism-oriented framework from sampling to system execution. For continuous models, we organize acceleration into five complementary categories by target cost: (i)~sampling and trajectory optimization (solvers, step distillation, and trajectory reconstruction) to reduce evaluations; (ii)~efficient attention and sparsity (structured/adaptive masks and linear/hybrid operators), (iii)~caching and computation reuse (cross-step features, historical KV, and cross-request reuse), and (iv)~model compression and low-precision computation (quantization, pruning, and compact latents) to lower per-step cost; and (v)~system optimization and parallel execution (IO-aware kernels, offloading, distributed parallelism, and serving/streaming schedules) to minimize exposed memory, communication, and scheduling overhead. Across categories, we analyze work saved, adaptation requirements, and trade-offs among quality, latency, and memory. We extend this analysis to diffusion language models (dLLMs), in which bidirectional attention allows the representations of committed tokens to evolve across rounds. We examine parallel commitment, selective state reuse, speculative verification, and specialized serving. Across tasks, structural dependencies determine which approximations preserve properties such as temporal coherence in video and geometric fidelity in 3D. The pipeline's critical path, including decoding and communication, determines the end-to-end gains achieved in practice. Accordingly, rigorous evaluation requires stated baselines, matched workloads, explicit timing boundaries, and task-specific quality criteria. Finally, we analyze error interactions under composed optimizations and outline open challenges in joint computation allocation, cross-task transfer, hardware-aware serving, and reliable parallel progress for long-horizon, real-time, and multimodal generation.

\end{unicomabstract}
\unicomkeywords{Diffusion Models, Efficient Generation, Accelerated Sampling,
  Cross-Task Analysis, Diffusion Language Models.}

\clearpage
\section{Introduction}
\label{sec:introduction}
\subsection{Development of Diffusion Models and Efficiency Bottlenecks}
\label{sec:intro:bottlenecks}

Diffusion models formulate generation as a sequence of learned refinement steps, transforming a noisy state into a structured sample. Denoising diffusion probabilistic models established this approach for image synthesis~\cite{ho2020ddpm}, while latent diffusion reduced the dimensionality of the space in which refinement is performed~\cite{rombach2022highresolutionimagesynthesislatent}. Transformer-based denoisers further extended the architectural design space~\cite{peebles2023scalablediffusionmodelstransformers}. These developments make efficiency a question about both the generation procedure and the model that executes each step.

The computational burden has several sources. Repeated model evaluations accumulate along the sampling trajectory; high-resolution or long-duration outputs increase the amount of data processed at each evaluation; and intermediate activations, attention operations, and data movement place pressure on memory and hardware utilization. Reducing the number of sampling steps addresses only part of this burden. A shorter trajectory can still be expensive if each evaluation processes a large sequence, and a lower operation count may produce little latency improvement when memory access or communication dominates execution.

The balance between these costs varies across tasks. Image generation must preserve spatial detail and prompt alignment, whereas video generation must also maintain motion and temporal consistency over longer contexts~\cite{https://doi.org/10.48550/arxiv.2204.03458}. Audio generation introduces continuity constraints over time, and 3D generation requires consistency across the underlying geometry and its rendered views~\cite{liu2023audioldm,jun2023shapegeneratingconditional3d}. Diffusion world models add a further requirement: predictions condition subsequent interaction and must support coherent environment dynamics~\cite{alonso2024diffusionworldmodelingvisual}. These differences motivate a review that connects acceleration mechanisms to task-specific dependencies and quality constraints.

We therefore treat efficient generation as a constrained design problem: reduce latency, memory use, or resource consumption while retaining the properties required by the application. This perspective distinguishes an algorithmic reduction in work from a deployment-level improvement, and makes the quality cost of approximation part of the comparison.

\begin{figure}[h]
\centering
\includegraphics[width=\linewidth]{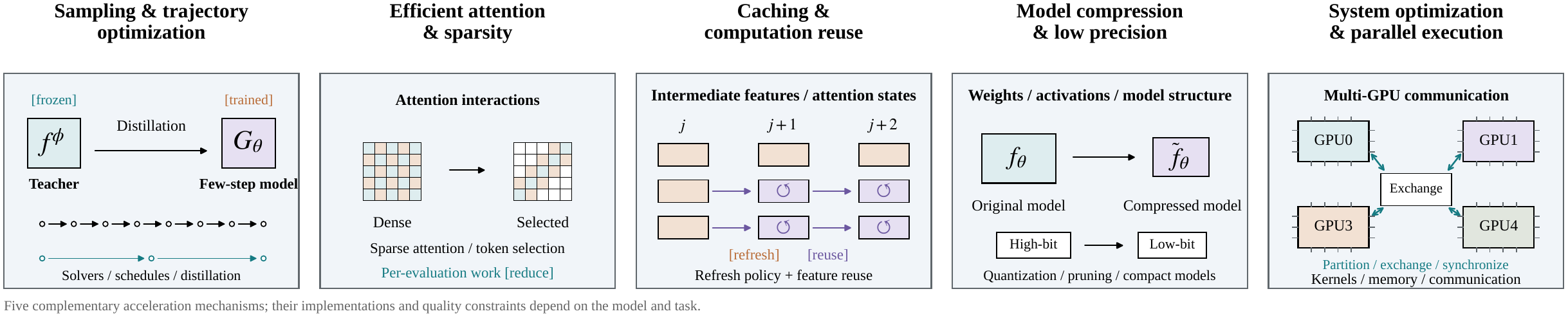}
\caption{Five complementary mechanisms for diffusion inference acceleration. From left to right: sampling and trajectory optimization reduces network evaluations through solvers, schedules or distillation; efficient attention and sparsity reduce interactions and active computation; caching reuses intermediate states under a refresh policy; model compression reduces structural or numerical costs; and system optimization coordinates execution and multi-GPU communication. The exchange block denotes logical communication, not a dedicated central device. The schematics illustrate mechanisms rather than measured speedups; preparation requirements and quality trade-offs depend on the specific method.}
\label{fig:acceleration_overview}
\end{figure}

Figure~\ref{fig:acceleration_overview} summarizes this design space through five complementary mechanisms: changing the sampling trajectory, reducing per-evaluation work, reusing intermediate states, compressing models and numerical representations, and reorganizing execution. These mechanisms are shared across generation tasks, while their applicability depends on model architecture, state dependencies, quality requirements and available hardware.

\subsection{Related Surveys and Scope of This Review}
\label{sec:intro:scope}

Existing reviews provide complementary views of diffusion efficiency. The cache-focused survey by Liu et al. organizes methods into static and dynamic caching and examines their applications across generation tasks~\cite{liu2025cachesurvey}. \emph{Efficient Diffusion Models: A Comprehensive Survey from Principles to Practices} discusses efficiency across multiple stages of model development and deployment~\cite{Ma_2025}. The video-focused review by Shao et al. organizes acceleration around step distillation, efficient attention, model compression, and cache/trajectory optimization, with particular attention to video deployment~\cite{shao2026efficientvideodiffusionmodels}. Table~\ref{tab:survey_comparison} summarizes the scope and organizing perspective of these reviews. The present review adopts a mechanism-oriented perspective across tasks and separates system execution from the model and sampling modifications with which it is often combined.

\begin{table}[!htbp]
\centering
\small
\caption{Scope and emphasis of related diffusion efficiency surveys.}
\label{tab:survey_comparison}
\setlength{\tabcolsep}{5pt}
\renewcommand{\arraystretch}{1.12}
\begin{tabular*}{\linewidth}{@{\extracolsep{\fill}}lll@{}}
\toprule
\textbf{Survey} & \textbf{Scope} & \textbf{Emphasis} \\
\midrule
Cache survey~\citep{liu2025cachesurvey} & Cross-task caching & Static and dynamic cache policies \\
Efficient diffusion~\citep{Ma_2025} & Model lifecycle & Model design, training, inference, deployment \\
Video diffusion~\citep{shao2026efficientvideodiffusionmodels} & Video inference & Distillation, attention, compression, reuse \\
\textbf{This review} & Cross-task inference & Five cost targets, transfer limits, dLLMs \\
\bottomrule
\end{tabular*}
\end{table}

Our main scope covers inference acceleration for continuous diffusion and related flow-based generation, with visual generation as the main body and audio, 3D, world-model, and joint-multimodal tasks providing cross-task comparisons. Flow matching is included because its iterative neural evaluation shares important computational questions with diffusion sampling, although the formulations and training objectives are distinct~\cite{lipman2023flowmatchinggenerativemodeling}. Training procedures are discussed when they create a faster generator or change the cost of inference. Diffusion language models (dLLMs) are examined separately in Section~\ref{sec:dllm}. Building on the collected literature~\cite{lin2026efficientdllm}, that section selects methods that reveal the consequences of discrete commitment, parallel prediction, and bidirectional state reuse, rather than reproducing a complete dLLM taxonomy.

Throughout the review, each method is examined through the same sequence of questions: which cost does it target, what mechanism removes or reorganizes that cost, what additional training or implementation is required, and under which conditions is the resulting efficiency gain useful? Cross-task comparison then asks whether those conditions survive a change in output structure or deployment setting.

\subsection{Contributions and Organization}
\label{sec:intro:contributions}

This review makes three connected contributions:
\begin{itemize}
\item \textbf{A mechanism-oriented cross-task framework.} Five categories link acceleration to sampling and trajectory design, attention and sparse computation, caching and reuse, compression and numerical precision, and system execution. Their comparison identifies the costs reduced and the task conditions under which those reductions are useful.
\item \textbf{A focused account of efficient dLLM inference.} Representative methods are organized around token decisions, state reuse, and speculative verification or AR collaboration. This perspective makes the special requirements of discrete generation explicit within a broader review of diffusion acceleration.
\item \textbf{An analysis of transfer and its limits.} We compare continuous and discrete generation through evolving states, iteration dependencies, redundant computation, and quality control. Shared principles are connected to the mechanisms that require redesign when the task or generation paradigm changes.
\end{itemize}

Section~\ref{sec:preliminaries} introduces the foundations, cost definitions, categories, and evaluation criteria. Section~\ref{sec:landscape} examines the research landscape and emerging questions across tasks. Sections~\ref{sec:sampling}--\ref{sec:systems} review the five acceleration categories. Section~\ref{sec:dllm} focuses on efficient diffusion language inference, followed by open problems in Section~\ref{sec:future} and a synthesis in Section~\ref{sec:conclusion}.

\section{Preliminaries and Categorization}
\label{sec:preliminaries}
\subsection{Diffusion Processes and Flow Matching}
\label{sec:prelim:process}

Let $x_0$ denote a clean sample and $c$ its conditioning information. In a latent-space model, an encoder maps the sample to $z_0=E(x_0)$, and an output decoder reconstructs the generated sample after sampling. A common Gaussian perturbation representation is
\begin{equation}
 z_t=\alpha_t z_0+\sigma_t\epsilon,
 \qquad \epsilon\sim\mathcal{N}(0,I),
 \label{eq:perturbation}
\end{equation}
where $t$ indexes noise level and $\alpha_t$ and $\sigma_t$ specify the signal and noise schedules. A neural denoiser estimates a quantity such as the clean state or added noise, and a sampler uses this estimate to update the current state. Pixel-space models follow the same notation by taking $z_0=x_0$. The representation and sampler jointly determine the amount and organization of inference work~\cite{ho2020ddpm,rombach2022highresolutionimagesynthesislatent}.

Flow matching instead learns a vector field along a prescribed probability path. Let $(z_{\mathrm{s}},z_{\mathrm{d}},c)$ be drawn from the endpoint and conditioning distribution, let $t$ be sampled over the path, and let $z_t$ be drawn from the conditional path given the endpoints. A representative objective is
\begin{equation}
 \mathcal{L}_{\mathrm{FM}}
 =\mathbb{E}_{z_{\mathrm{s}},z_{\mathrm{d}},c,t,z_t}\!\left[
 \left\|v_\theta(z_t,t,c)-u_t(z_t\mid z_{\mathrm{s}},z_{\mathrm{d}})\right\|_2^2
 \right],
 \label{eq:flowmatching}
\end{equation}
where $u_t$ is the conditional target velocity and $v_\theta$ is the learned field. Generation integrates the learned field from the source distribution toward data~\cite{lipman2023flowmatchinggenerativemodeling}. The direction assigned to $t$ is a convention: the perturbation in Equation~\ref{eq:perturbation} and a source-to-data flow need not use the same orientation. Their common efficiency concern is repeated neural evaluation along a numerical trajectory, rather than an identical stochastic process.

We distinguish the number of sampler updates $S$ from the number of function evaluations (NFE), denoted by $K$. A solver can evaluate the model more than once per update or reuse earlier estimates. Guidance can also require conditional and unconditional predictions, whether executed separately or batched together. Consequently, NFE must be reported with its counting convention and cannot be inferred from the update count alone. DDIM and DPM-Solver illustrate how sampling design can change the evaluation budget without retraining the underlying denoiser~\cite{song2020ddim,lu2022dpmsolver}.

\subsection{Model Architectures, Task Scope, and Computational Costs}
\label{sec:prelim:costs}

Architectures determine where each evaluation spends its resources. U-Net denoisers combine multiscale feature processing with skip connections, while diffusion transformers (DiTs) process spatial or spatiotemporal tokens through repeated transformer blocks~\cite{peebles2023scalablediffusionmodelstransformers}. For a dense attention layer with $N$ tokens and head dimension $d$, attention interactions involve an $\mathcal{O}(N^2d)$ term. This does not imply that attention always dominates a complete denoiser: projections, feed-forward layers, convolutions, and memory movement may remain substantial. Likewise, an exact attention kernel can reduce memory traffic without changing the dense attention pattern or its arithmetic scaling~\cite{dao2022flashattention}.

For a specified sequential workload, a useful latency decomposition is
\begin{equation}
 T_{\mathrm{total}}
 =T_{\mathrm{condition}}+\sum_{j=1}^{K}\tau_j
 +T_{\mathrm{decode}}+T_{\mathrm{exposed}},
 \label{eq:latency}
\end{equation}
where $\tau_j$ is the time attributed to the $j$th model evaluation, including its local computation and memory access. $T_{\mathrm{exposed}}$ accounts only for additional execution, transfer, or coordination time not already included in those measurements. The terms depend on model, precision, token count, batch size, and hardware. Under asynchronous execution, overlapping activities must be measured on the critical path rather than added twice. Equation~\ref{eq:latency} is therefore an accounting framework, not a universal predictor of speedup.

Memory provides a second cost axis. At peak execution, device memory holds parameters, active intermediate tensors, cached states, and implementation workspaces. Their relative sizes depend on the execution schedule: reducing parameters does not necessarily resolve an activation bottleneck, while a cache that saves arithmetic can increase persistent storage. We therefore treat latency and peak memory as separate objectives rather than using either as a proxy for the other.

Table~\ref{tab:taskcosts} summarizes the task characteristics used in this review. These are analytical considerations for comparing methods, not measurements showing that a particular bottleneck dominates every model in a task. A world model may generate video, and a multimodal generator may emit image or audio outputs; these labels can overlap. Joint multimodal generation here refers to explicit coordination between multiple modalities or output streams; ordinary text conditioning does not automatically place every text-to-image method in that category.

\begin{table}[!htbp]
\centering
\small
\caption{Computational and quality considerations by generation task.}
\label{tab:taskcosts}
\setlength{\tabcolsep}{5pt}
\renewcommand{\arraystretch}{1.12}
\begin{tabular*}{\linewidth}{@{\extracolsep{\fill}}lll@{}}
\toprule
\textbf{Task} & \textbf{Main cost sources} & \textbf{Quality constraints} \\
\midrule
Image & Resolution, latent size, denoiser calls & Detail, layout, condition alignment \\
Video & Space--time context, memory, communication & Motion, temporal consistency, drift \\
Audio & Duration, temporal context, decoding & Continuity, fidelity, alignment \\
3D & Representation size, geometry, rendering & Topology, shape, view consistency \\
World modeling & Rollout horizon, recurring inference & State consistency, useful dynamics \\
Joint multimodal & Stream processing, fusion & Synchronization, semantic consistency \\
\bottomrule
\end{tabular*}
\end{table}

\subsection{Categorization of Acceleration Methods}
\label{sec:prelim:taxonomy}

We classify a method by the principal mechanism through which it changes inference cost. The categories are complementary: a deployed pipeline may combine several of them, and a paper can receive secondary labels when it introduces multiple mechanisms. Figure~\ref{fig:acceleration_overview} provides a conceptual overview of these mechanisms, while Table~\ref{tab:categories} connects the classification to Equation~\ref{eq:latency}. Trajectory methods primarily change the evaluation budget or the states visited; sparse computation, reuse, and compression primarily change the cost of an evaluation; system methods change how these operations and their overheads appear on the execution path. This mapping describes the primary effect, since combinations can alter more than one term.

\paragraph{Sampling steps and generation trajectories.}
These methods change the evaluation budget or the path followed by the generator. Training-free solvers modify numerical integration, whereas distillation trains a generator to reproduce a more expensive sampling procedure with fewer evaluations~\cite{lu2022dpmsolver,salimans2022progressivedistillationfastsampling}. They share a cost target but differ in training requirements and approximation. A distillation method whose main result is a shorter trajectory belongs here; distillation into a smaller network can additionally receive a compression label.

\paragraph{Efficient attention and sparse computation.}
These methods reduce the work performed within an evaluation by changing attention interactions, selecting tokens, or exploiting a more economical attention formulation. Their effectiveness depends on whether the removed interactions are unnecessary for the task and whether the retained pattern maps efficiently to hardware. Exact IO-aware kernels are discussed primarily under systems, with cross-references here, because they change execution rather than remove attention dependencies~\cite{dao2022flashattention}.

\paragraph{Caching and computation reuse.}
These methods replace selected recomputation with stored or predicted intermediate states. DeepCache, for example, uses redundancy across denoising steps to reuse high-level U-Net features while updating lower-level features~\cite{ma2023deepcache}. A cache policy must specify what is stored, when it is refreshed, and whether reuse is exact or approximate. Unchanged inputs do not by themselves guarantee unchanged hidden states when the surrounding generated context evolves.

\paragraph{Model compression and low precision.}
These methods reduce parameter, activation, or representation costs through lower precision or structural compression. Their utility depends on numerical sensitivity and available execution kernels. Q-Diffusion demonstrates why calibration must account for the multi-timestep behavior of a diffusion network~\cite{li2023qdiffusionquantizingdiffusionmodels}. Lower bitwidth does not automatically yield proportional latency reduction. We distinguish persistent model or representation compression from inference-time sparse selection, while recording overlapping mechanisms explicitly.

\paragraph{System optimization and parallel execution.}
These methods reorganize kernels, memory, work placement, and scheduling. Parallelism can distribute a single generation workload, overlap pipeline stages, or increase throughput across requests. DistriFusion combines patch distribution with reuse of previous-step context to support asynchronous communication~\cite{li2024distrifusion}; it therefore spans both systems and reuse. This example also shows why distributed execution must be characterized by its numerical assumptions and hardware budget, rather than treated as an automatically exact acceleration.

\begin{table}[!htbp]
\centering
\small
\caption{Primary cost targets and trade-offs of the five acceleration mechanisms.}
\label{tab:categories}
\setlength{\tabcolsep}{5pt}
\renewcommand{\arraystretch}{1.12}
\begin{tabular*}{\linewidth}{@{\extracolsep{\fill}}lll@{}}
\toprule
\textbf{Mechanism} & \textbf{Cost target} & \textbf{Main trade-off} \\
\midrule
Sampling and trajectories & Evaluations and trajectory & Quality, training or calibration \\
Attention and sparsity & Token interactions & Dependency coverage, routing cost \\
Caching and reuse & Repeated state computation & State drift, refresh overhead \\
Compression and precision & Arithmetic, model or latent size & Fidelity, hardware support \\
Systems and parallelism & Execution and data movement & Coordination, resource cost \\
\bottomrule
\end{tabular*}
\end{table}

\subsection{Efficiency and Quality Evaluation}
\label{sec:prelim:evaluation}

Efficiency comparisons should report both the amount of work and the time or resources needed to execute it. NFE and operation counts help explain algorithmic changes, whereas latency, throughput, and peak memory describe their realization on a specified platform. For real-time or streaming generation, first-output delay and steady-state throughput answer different questions. Distributed results additionally require device count and communication conditions; shorter single-sample latency with more devices is not evidence of improved resource efficiency by itself.

A reported speedup is meaningful only relative to a stated baseline. We use $\eta=T_{\mathrm{base}}/T_{\mathrm{method}}$ for matched timing boundaries, and retain the model checkpoint, input/output size, sampler, precision, batch size, hardware, and implementation alongside each value. Measurements limited to a denoiser must be distinguished from end-to-end measurements that include conditioning and decoding. If an unchanged component consumes fraction $p$ of baseline latency and the remaining component is accelerated by factor $r$, the ideal sequential end-to-end gain is $1/[p+(1-p)/r]$. This relation explains why strong kernel or denoiser gains can have a smaller application-level effect; it assumes fixed workload and no new overhead.

Quality must be evaluated in the terms of the task. Distributional and perceptual image measures do not fully capture temporal coherence, audio intelligibility, geometric validity, or interactive state consistency. We therefore retain task-appropriate quality metrics and, where available, condition alignment and human evaluation. An approximate method should be compared at an explicitly stated quality tolerance rather than described as universally lossless. For methods requiring calibration, fine-tuning, or distillation, the adaptation cost is recorded separately from inference performance.

\section{Research Landscape and Emerging Trends}
\label{sec:landscape}

\begin{figure}[!htbp]
    \centering
    \includegraphics[width=\linewidth]{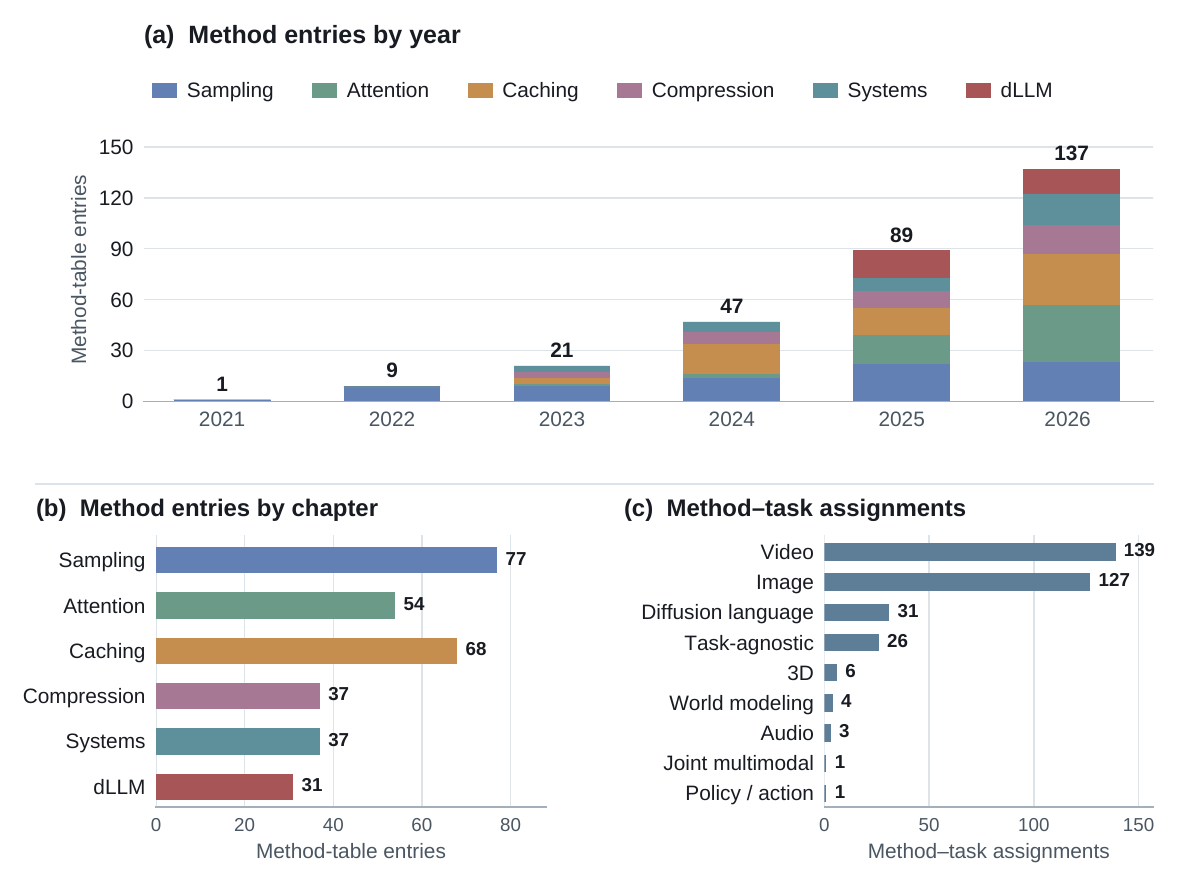}
    \caption{Coverage of the method inventories\ifsurveyappendices{} in Appendices~\ref{app:methods:sampling}--\ref{app:methods:dllm}, corresponding to\else{} for\fi{} Sections~\ref{sec:sampling}--\ref{sec:dllm}. (a) Annual totals of method-table entries, grouped by chapter. (b) Method-table entries by chapter. (c) Method--task assignments across nine task labels. Entries associated with multiple tasks contribute to each applicable label.}
    \label{fig:literature-statistics}
\end{figure}

The research landscape is shaped by two changes: diffusion generators operate on increasingly demanding output structures, and acceleration extends from shortening the sampling trajectory to reducing the cost of each evaluation and its execution. We examine these changes through representative work and the task distinctions defined in Section~\ref{sec:preliminaries}. Figure~\ref{fig:literature-statistics} summarizes the method inventories for Sections~\ref{sec:sampling}--\ref{sec:dllm}, and Figure~\ref{fig:mechanism-task-sankey} shows the task coverage of the methods in Sections~\ref{sec:sampling}--\ref{sec:systems}.

\paragraph{Literature selection and coding.}
The inventory\ifsurveyappendices{} in Appendices~\ref{app:methods:sampling}--\ref{app:methods:dllm}\fi{}
draws on related surveys and the primary papers cited in Sections~4--9.
It includes work that changes the cost or execution of diffusion, flow-based,
or masked diffusion language inference, including training or adaptation
that enables faster inference. Background papers are used to describe
the tasks and models. We assign each entry to a primary mechanism within
its chapter and take its year from the bibliography. Task labels follow
the reported evaluation setting and, where needed, the cited title.
Entries without a specific task are coded as task-agnostic.
Cross-listed methods count once per listed row and once per citation key
in the unique-key total. The inventory reflects the scope of this review.

\subsection{Research across Generation Tasks}
\label{sec:landscape:tasks}

\paragraph{Images and video.}
Image generation provides representative examples across several mechanisms. DDIM and DPM-Solver change sampling trajectories, progressive distillation learns shorter ones, DeepCache reuses intermediate features, and Q-Diffusion adapts numerical precision~\cite{song2020ddim,lu2022dpmsolver,salimans2022progressivedistillationfastsampling,ma2023deepcache,li2023qdiffusionquantizingdiffusionmodels}. These methods establish complementary controls over inference cost. Video extends generation along time~\cite{https://doi.org/10.48550/arxiv.2204.03458}, making efficiency depend on both spatial detail and temporal dependencies. The video-focused survey~\cite{shao2026efficientvideodiffusionmodels} accordingly brings attention, memory, and repeated denoising into the same discussion. A useful cross-task question is whether a mechanism that preserves individual frames also preserves their motion and consistency over longer outputs.

\paragraph{Audio and 3D.}
AudioLDM demonstrates latent diffusion for text-to-audio generation, while Shap-E applies conditional diffusion to encoded 3D implicit functions~\cite{liu2023audioldm,jun2023shapegeneratingconditional3d}. These examples broaden the representations on which iterative generation operates; they are task foundations, not entries automatically counted as acceleration methods. For audio, shorter sampling and temporal reuse must preserve continuity and condition alignment. For 3D, the representation determines whether cost lies in token processing, geometry decoding, or rendering. A similar NFE reduction can therefore produce different application-level gains across these pipelines.

\paragraph{World models and joint multimodal generation.}
DIAMOND uses diffusion within a learned environment model, connecting visual fidelity to the usefulness of generated dynamics~\cite{alonso2024diffusionworldmodelingvisual}. This setting adds feedback: an error can affect subsequent interaction rather than only a completed sample. Joint multimodal generation similarly requires multiple streams to remain aligned. In both cases, evaluating isolated outputs is insufficient; timing and consistency must be assessed along the generated or interactive sequence. World-model and video labels can overlap, so task coverage should be recorded with multiple labels rather than forced into disjoint bins.

These contrasts motivate a common unit of comparison: the mechanism, the state or computation it changes, and the dependency it must preserve. They also explain why the number of papers collected for a task cannot by itself establish either its maturity or its remaining opportunities.

\subsection{Evolution of Acceleration Mechanisms}
\label{sec:landscape:temporal}

\paragraph{From fewer evaluations to cheaper evaluations.}
DDIM makes alternative sampling trajectories central to efficient generation, while DPM-Solver develops specialized numerical integration and progressive distillation transfers a more expensive sampler into a shorter learned procedure~\cite{song2020ddim,lu2022dpmsolver,salimans2022progressivedistillationfastsampling}. The distinction is methodological: changing the solver and changing the trained generator can target the same evaluation budget with different adaptation costs. These approaches remain useful as components of later pipelines rather than forming a sequence in which one eliminates the others.

The broader design space also addresses the work repeated within a trajectory. Q-Diffusion accounts for changing activation distributions when reducing precision, and DeepCache exploits redundancy in intermediate representations~\cite{li2023qdiffusionquantizingdiffusionmodels,ma2023deepcache}. Attention and token selection offer another control over per-step cost, particularly when outputs become long sequences~\cite{shao2026efficientvideodiffusionmodels}. The relevant trend is a broadening of the optimization target: from how often the network is evaluated to which parts are evaluated and how they are represented.

\paragraph{From isolated algorithms to coordinated execution.}
DistriFusion combines distributed work placement with previous-step context to reduce exposed communication~\cite{li2024distrifusion}. Its mechanism spans execution and reuse, illustrating why algorithm and system design increasingly need to be considered together. A shorter trajectory changes the opportunity for caching; sparsity changes tensor sizes and communication; low precision changes both arithmetic and transfer cost. This motivates tracking combinations and their interactions rather than multiplying independent maximum speedups.

Figure~\ref{fig:literature-statistics} summarizes 304 method-table entries\ifsurveyappendices{} across Appendices~\ref{app:methods:sampling}--\ref{app:methods:dllm}\fi{}, corresponding to 303 distinct citation keys. Each table entry counts once in the chapter and year totals. Task coverage is multi-label: an entry associated with several tasks contributes one assignment to each, giving 338 method--task assignments across nine task labels. Video and image account for 139 and 127 assignments, respectively.


\subsection{Mechanism--Task Relationships and Emerging Questions}
\label{sec:landscape:hotspots}

A mechanism--task view asks where an optimization is useful and which constraints limit transfer. Step reduction is relevant wherever repeated evaluations dominate, but difficult temporal or geometric dependencies may require a different quality tolerance. Sparse attention becomes attractive as context grows, yet the right interaction pattern depends on motion, long-range audio structure, or geometric correspondence. Reuse requires stable states and a refresh policy matched to the task. Compression depends on numerical sensitivity, whereas parallel execution depends on the available hardware and communication path. These are relationships to test, not a measured ranking of task hotspots.

\begin{figure}[!htbp]
    \centering
    \includegraphics[width=\linewidth]{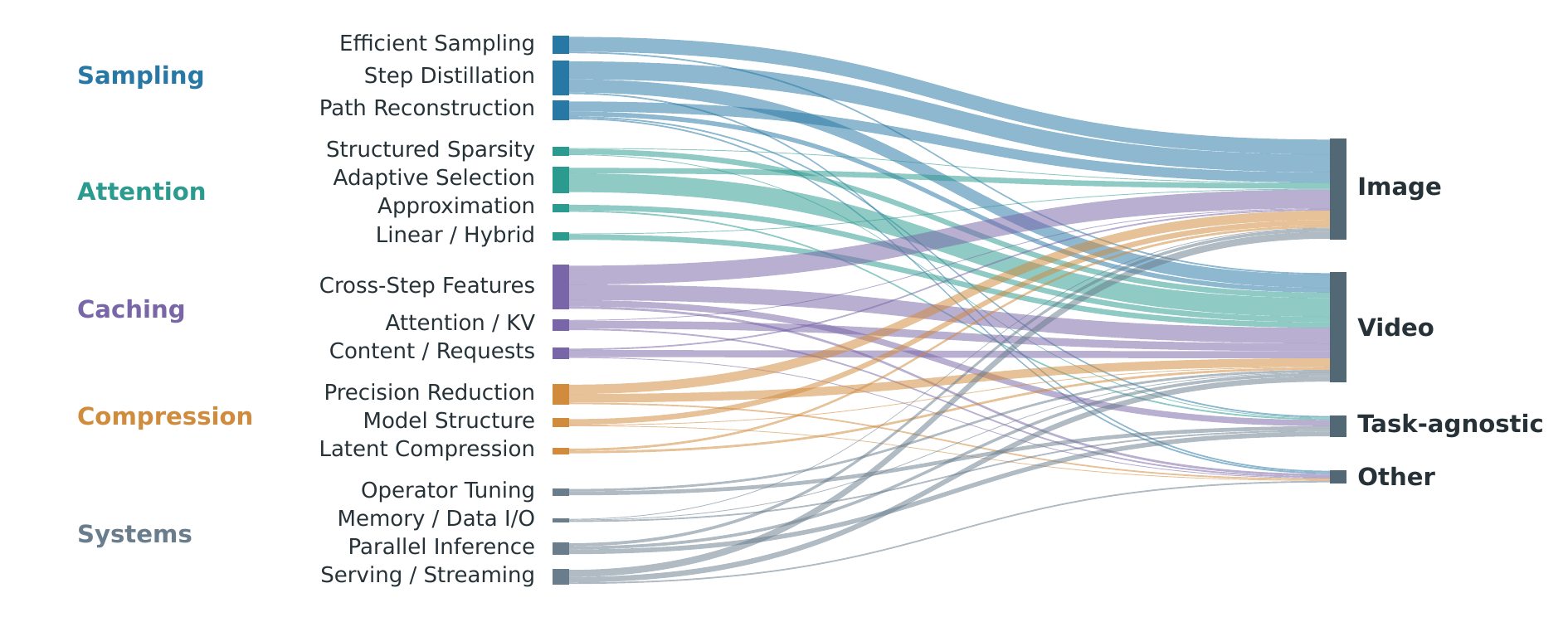}
    \caption{Task coverage of 17 acceleration method subcategories in Sections~\ref{sec:sampling}--\ref{sec:systems}. Colors identify the five sections, and ribbon widths are proportional to the number of method--task assignments (307 in total). Other combines five task categories with fewer than 10 assignments each.}
    \label{fig:mechanism-task-sankey}
\end{figure}

Figure~\ref{fig:mechanism-task-sankey} maps 17 method subcategories from Sections~\ref{sec:sampling}--\ref{sec:systems} to four task groups. It shows 307 of the 338 method--task assignments counted in Figure~\ref{fig:literature-statistics}(c), with the 31 diffusion-language assignments omitted. Each ribbon has a width proportional to its assignment count. The Other group combines 3D, world modeling, audio, joint multimodal generation, and policy or action generation, each with fewer than 10 assignments before grouping. The 15 combined assignments retain their links to the corresponding method subcategories.

Three questions follow for the technical chapters. First, \emph{what is the transferable unit}? A numerical solver may transfer across compatible vector-field formulations, whereas a U-Net feature cache depends on architectural state boundaries. Second, \emph{which dependency survives the approximation}? Local savings must be evaluated against temporal, geometric, or cross-modal consistency. Third, \emph{does reduced work reach the user}? Conditioning, decoding, synchronization, and scheduling can dominate after the denoiser becomes faster.

The same questions connect the continuous-generation body to dLLMs. Discrete token commitment and bidirectional representations introduce additional decisions that are not captured by simply assigning language methods to the five categories. Section~\ref{sec:dllm} therefore develops a focused analysis of those decisions, their reusable states, and their verification mechanisms.

\section{Sampling Steps and Generation Trajectory Optimization}
\label{sec:sampling}

Reverse generation in diffusion models repeatedly evaluates the denoising network, so the cost of a sample is governed by how many evaluations are performed and how expensive each one is. Denoising Diffusion Probabilistic Models (DDPM) formulate generation as a reverse Markov chain \cite{ho2020ddpm}; the score-based stochastic differential equation (SDE) framework extends the process to continuous time, where the probability-flow ODE (PF-ODE) turns sampling into an ODE initial-value problem \cite{song2020score}. For comparability, this chapter uses the total number of denoising-network evaluations (NFE) per sample as an algorithmic cost proxy,
\[
\mathrm{NFE}=\sum_{k=1}^{S} n_k,
\]
where $S$ is the nominal number of sampling steps and $n_k$ is the number of denoiser evaluations associated with step $k$. The approximate denoiser cost is $\sum_{k=1}^{S}\sum_{j=1}^{n_k} C_{k,j}$, where $C_{k,j}$ is the cost of evaluation $j$ at step $k$. A multistep integrator may reuse earlier evaluations, while a multistage rule may require more than one new evaluation per step. Classifier-free guidance (CFG) evaluates conditional and unconditional branches~\cite{ho2022classifier}; implementations can batch them into one invocation. We record the counting convention, distinguish full-backbone from auxiliary evaluations, and treat measured latency as a separate outcome.

\begin{figure}[!htbp]
    \centering
    \includegraphics[width=\linewidth]{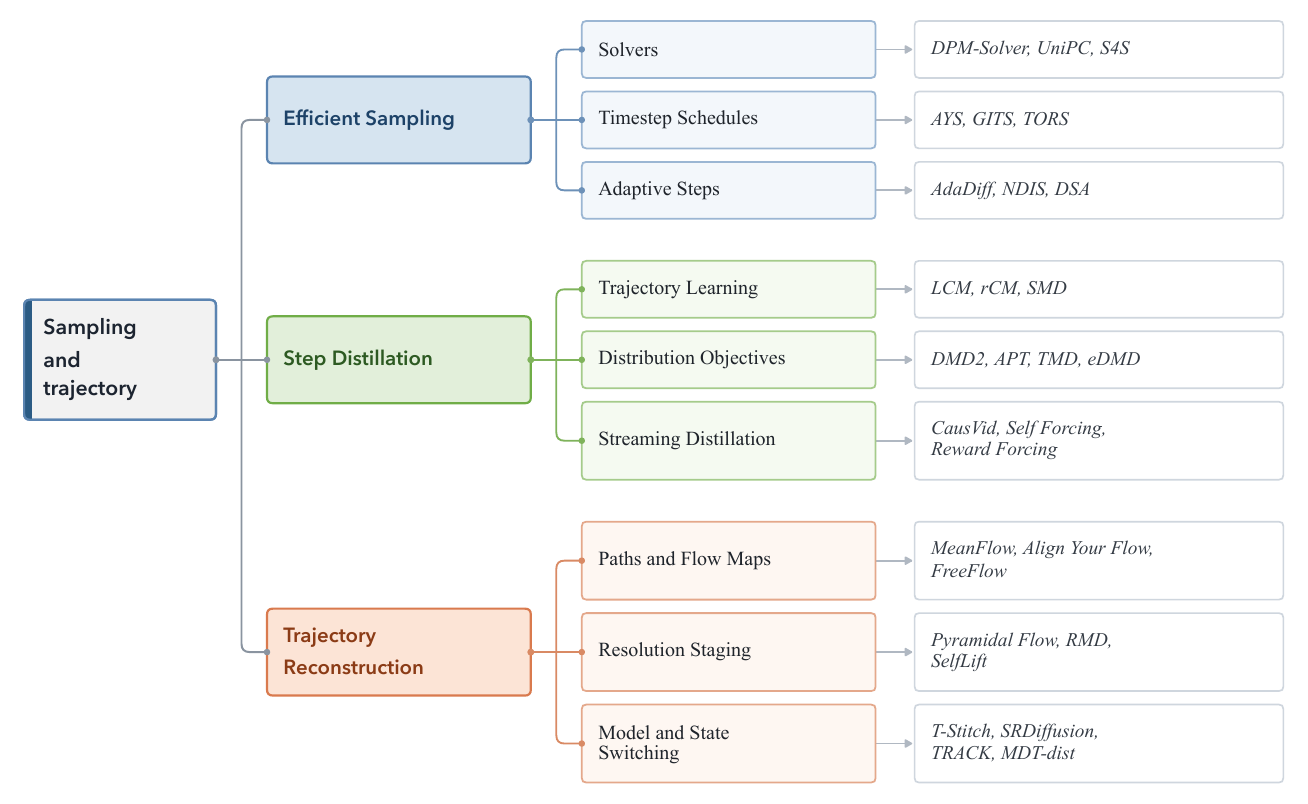}
    \caption{Taxonomy of sampling-step and trajectory acceleration. Branches follow the primary mechanism of each method. The leaves show examples\ifsurveyappendices\space from Table~\ref{tab:routes}\fi.}
    \label{fig:sampling-taxonomy}
\end{figure}

\begin{figure}[!htbp]
    \centering
    \includegraphics[width=\linewidth]{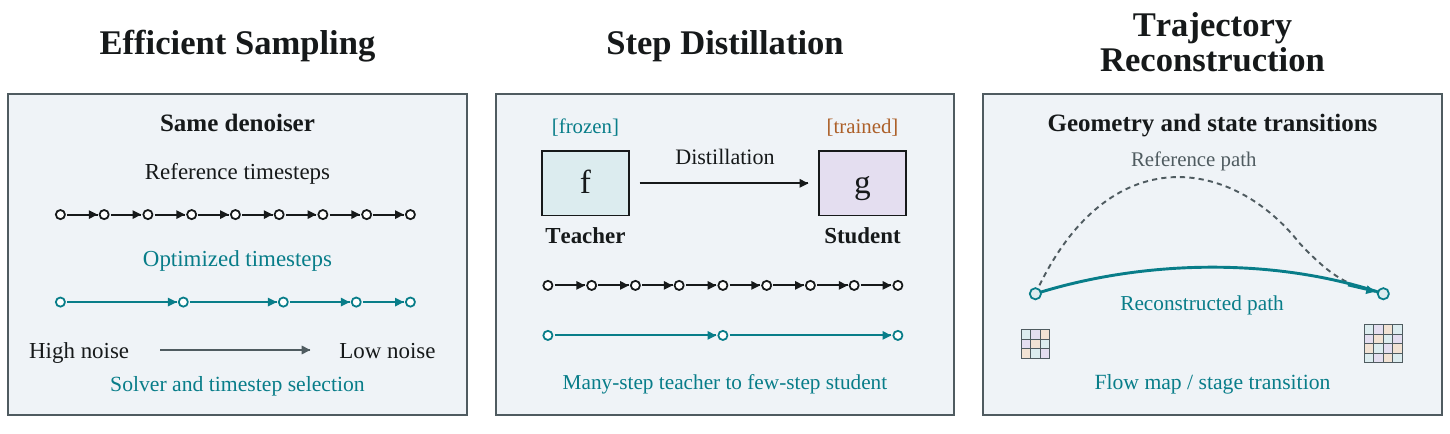}
    \caption{Sampling-step and trajectory acceleration. Efficient sampling changes the numerical update or timestep grid used with a denoiser. Step distillation trains a student to generate in fewer steps. Trajectory reconstruction changes the path, maps across time intervals, or switches the resolution, denoiser, or state representation across stages.}
    \label{fig:sampling-overview}
\end{figure}

\ifsurveyappendices Table~\ref{tab:routes} in Appendix~\ref{app:methods:sampling} compares the mechanisms, preparation requirements, and reported settings of the methods discussed in this section. \fi{}Figure~\ref{fig:sampling-taxonomy} groups them by their primary intervention. Efficient sampling changes the numerical update, timestep grid, or allocation of steps. Step distillation and related few-step training change the learning objective. Trajectory reconstruction reorganizes the path, resolution, or denoiser capacity across stages. These interventions can coexist within the same system. Figure~\ref{fig:sampling-overview} illustrates how these interventions change the sampling grid, transfer the generation process to a few-step student, or reshape the trajectory.


\subsection{Efficient Sampling}
\label{sec:sampling-efficient}

These methods target fewer evaluations or lower discretization error at a fixed budget. Most retain the denoising backbone. Some fit a solver, search a schedule, or train a budget controller.

\subsubsection{Solvers}
\label{sec:sampling-solvers}

With the model and budget fixed, samplers differ in how they discretize the reverse process (integration order and step size), in the ODE parameterization (noise, data, or velocity prediction), and in whether randomness is injected along the trajectory.

DDIM derives a skippable deterministic reverse process from a non-Markovian forward process; sampling is deterministic when $\eta=0$, and the paper relates the continuous limit to a probability-flow ODE \cite{song2020ddim}. The reported 10--50$\times$ wall-clock speedup over 1000-step DDPM spans different operating points: on CIFAR-10, the Fr\'echet Inception Distance (FID) is 4.16 at 100 steps but rises to 6.84 at 20. Pseudo Numerical Methods for Diffusion Models (PNDM) split the update into a higher-order gradient estimate and a nonlinear transfer that keeps the iterate on the data manifold, while the Diffusion Exponential Integrator Sampler (DEIS) builds an exponential integrator from the semilinear structure of the reverse process. On CIFAR-10, the fast variant F-PNDM reaches FID 3.95 at 50 steps, close to the re-evaluated 1000-step DDIM baseline (DDIM$^*$, 4.00) in the same table, a 20$\times$ reduction in the number of steps \cite{liu2022pseudo,zhang2022fast}. All three operate without retraining. DDIM uses a reverse process defined on a shorter timestep subsequence. PNDM and DEIS modify the numerical update to improve integration at coarse step sizes.

DPM-Solver integrates the linear term analytically and applies a high-order approximation only to the nonlinear term, reaching FID 4.70/2.87 at 10/20 NFE on CIFAR-10 with a reported 4--16$\times$ reduction in NFE relative to the best previous training-free sampler \cite{lu2022dpmsolver}. DPM-Solver++ uses data prediction and thresholding to stabilize high orders under strong guidance and requires only about 15--20 steps in that regime \cite{lu2025dpm}; UniPC unifies orders and parameterizations in a predictor--corrector form, reporting FID 3.87 (CIFAR-10) and 7.51 (ImageNet 256$\times$256) at 10 NFE with runtime and memory comparable to DPM-Solver++ \cite{zhao2023unipc}. EDM instead organizes preconditioning, the training noise distribution, the solver, and stochasticity into one design space and supplies a second-order Heun sampler; the solver transfers across models, but preconditioning and noise distribution usually need recalibration or retraining \cite{karras2022elucidating}. Restart injects controlled noise at restart points to contract accumulated discretization and approximation error, reporting roughly 10$\times$ and 2$\times$ fewer NFE on CIFAR-10 and ImageNet-64 at equal FID \cite{xu2023restart}. DPM-Solver-v3 and SA-Solver pursue the same goal through empirical-statistic parameterization and variance-controlled stochastic multistep updates \cite{zheng2023dpm,xue2023sa}. A practical implication is that solver order alone does not determine few-step quality; order, guidance strength, and step count are coupled and must be selected jointly.

Learned solvers extend this design space at very low budgets. AMED estimates a mean direction over a large ODE interval, while S4S fits the coefficients of a coarse solver to a stronger solver's outputs~\cite{Zhou_2024_CVPR,pmlr-v267-frankel25a}. S4S can alternate solver fitting with timestep search, and its five-NFE setting reports a CIFAR-10 FID of 3.73 in the paper's setup. Both retain the pretrained denoiser and incur offline fitting. LTC-Accel instead approximates a local transition from its coherence with adjacent steps~\cite{zhu2025accelerating}. Its gain depends on the accuracy of the revised update at the chosen step size.

\subsubsection{Timestep Schedules}
\label{sec:sampling-schedules}

Once the solver is fixed, the discrete noise levels determine where evaluation effort is spent. Align Your Steps (AYS) optimizes the timestep grid for a given model and solver by minimizing an upper bound on a path-space discrepancy~\cite{sabour2024align}. F-scheduler jointly selects the timestep grid and the steps at which FreeU is enabled~\cite{li2025f,si2024freeu}. Comparisons that also change the solver, guidance scale, or resolution confound the effect of the grid. Evaluating the schedule alone calls for holding these settings fixed.

GITS searches for a timestep grid by minimizing a trajectory-error objective under a fixed sampling budget~\cite{Xue_2024_CVPR}. TORS uses the geometry of reference trajectories to place timesteps at approximately equal increments of accumulated rotation, and evaluates 10-step image generation on FLUX.1-dev and Stable Diffusion 3.5~\cite{zhou2026tors}. AYS, GITS, and TORS choose a reusable grid for a given setting. Their setup costs and transfer across samplers are therefore distinct from per-sample prediction.

\subsubsection{Adaptive Steps}
\label{sec:sampling-adaptive}

An adaptive policy can change either the positions of the timesteps or the number of evaluations. AdaDiff predicts a request-level budget from prompt complexity~\cite{zhang2025adadiff}. Instance-aware discretization (NDIS) chooses a sample-specific grid and can improve quality at the same budget. Its CIFAR-10 comparison reports FID 3.31 at five NFE~\cite{Yuan_2026_CVPR}. Dynamic Step Allocation (DSA) learns a confidence head jointly with a few-step video generator and stops denoising individual frames at different budgets~\cite{le2026dsa}. Confidence can understate errors in difficult frames, so the distribution of allocated steps and the effect on temporal consistency matter alongside the mean budget.

The comparison across the three subtypes should hold the backbone, guidance, resolution, and nominal budget fixed where possible. A saved evaluation on an individual sample may still yield a smaller throughput improvement when a batch shares a common execution schedule.

\subsection{Step Distillation}
\label{sec:sampling-fewstep}

Few-step training directly optimizes the transitions used at small inference budgets. Some approaches transfer a teacher's trajectory or output distribution, while others learn a few-step mapping directly from data or a task objective. We group methods by the signal applied to the student and separate post-training cost from inference cost.

\subsubsection{Trajectory Learning}
\label{sec:sampling-consistency}

Trajectory-preserving methods either enforce pointwise correspondence along the teacher path or replace it with a self-consistency constraint. Progressive Distillation merges two adjacent steps per round; after compressing a teacher from 8192 steps, CIFAR-10 FID is 4.51/3.00/2.57 at 2/4/8 steps, while one step degrades to 9.12---repeated halving works at intermediate budgets but has a clear limit \cite{salimans2022progressivedistillationfastsampling}. Consistency Models instead learn a mapping from any noise level to the data endpoint, either by distillation or from scratch. On CIFAR-10, consistency distillation reaches FID 3.55/2.93 at 1/2 steps, against 8.34/5.58 for reproduced progressive distillation in the same table, 8.70/5.83 for from-scratch consistency training, and 2.04 for the EDM teacher \cite{song2023consistency}. The gap to the teacher shows that self-consistency does not fully replace integration. Latent Consistency Models (LCM) move CFG-augmented consistency distillation into latent space, reaching FID 11.10 at four steps on 512$\times$512 LAION-Aesthetic-6+ versus 15.12 for Guided-Distill; a 768$\times$768 2--4-step LCM costs about 32 A100 GPU-hours, and LCM-LoRA packages the result as an adapter that leaves the base weights unchanged \cite{luo2023latent,meng2023distillation,luo2023lcm}.

Later work generalizes the constraint, relaxes its boundary conditions, or extends it to larger models. Consistency Trajectory Models (CTM) extend self-consistency from a noise-level-to-endpoint mapping to arbitrary time pairs $(t,s)$, so one network supports both one-step generation and multi-step refinement, at the cost of constructing paired-time targets \cite{kim2024consistency}. Trajectory Consistency Distillation (TCD) relaxes the boundary condition with a semilinear consistency function; Phased Consistency Models (PCM) segment the trajectory and evaluate over 1--16 steps; Hyper-SD packs segmented consistency, human feedback, and score distillation into a single LoRA; Dual-Expert Consistency Model (DCM) assigns semantic and detail experts to alleviate conflicting gradients across time steps; and rCM scales continuous-time consistency to large models with a score regularizer \cite{zheng2024trajectory,wang2024phased,ren2024hyper,lv2025dual,zheng2026large}.

The state distribution on which targets are formed is another design choice. The SMD preprint samples states from the student's own rollout, regresses teacher velocity there, and uses a training-only residual branch to capture variation inside a coarse interval~\cite{zhang2026simplifying}. The paper evaluates four-NFE students across image, video, and 3D generation. This complements consistency constraints, which specify the relationship between predictions at different times, by changing where the regression targets are observed.

\subsubsection{Distribution Objectives}
\label{sec:sampling-distribution}

Pointwise constraints accumulate error when student capacity is limited, so distribution matching relaxes them to the output distribution, and adversarial training replaces an explicit distance with a discriminator. Distribution Matching Distillation (DMD) derives an approximate reverse-KL objective from the score difference between the target and student-generated distributions \cite{yinone}. DMD2 removes the regression loss, changes the training update schedule, and adds a GAN loss and multi-step training. On ImageNet-64, its Table~1 reports FID 1.51 at one step, or 1.28 with extended training. The EDM teacher scores 2.32 with ODE sampling and 1.36 with SDE sampling, both at 511 NFE \cite{yin2024improved}. Score Identity Distillation (SiD) avoids real data by expressing the Fisher divergence between the teacher-approximated real distribution and the student distribution through a score identity estimable from student samples, so its accuracy hinges on the teacher-score approximation error \cite{zhou2024score}.

Adversarial variants target fidelity and perceptual quality. Adversarial Diffusion Distillation (ADD) and SDXL-Lightning add adversarial losses in the one-to-four-step regime \cite{sauer2024adversarial,lin2024sdxl}; Adversarial Distribution Matching (ADM) aligns real and fake scores through a diffusion discriminator, and its DMDX pipeline (adversarial pretraining, then ADM fine-tuning) beats the DMD2 baseline for one-step SDXL, with multi-step ADM extended to SD3-Medium, SD3.5-Large, and CogVideoX \cite{lu2025adversarial}. LADD places discriminators on intermediate features of a frozen teacher: it re-noises generated latents and reads the full token sequence after each attention block, avoiding pixel decoding and the 518$\times$518 input limit of DINOv2 discriminators; four-step guidance-free sampling on Stable Diffusion 3 (8B) yields SD3-Turbo \cite{sauer2024fast}. Flash Diffusion combines an adversarial loss with denoising distillation and distribution matching, transfers to new backbones and tasks with a few GPU-hours of training, and composes with adapters and LoRAs \cite{chadebec2025flash}. The criterion shifts from trajectory consistency to distributional closeness, which improves fidelity at 1--4 steps but may reduce diversity.

Recent work refines the objective, restores coverage, and lowers training or deployment cost. Data-Forcing adds a teacher-score-discrepancy term to DMD2---the difference between real and student samples in the teacher's score field---and removes the GAN loss, improving diversity and fidelity on Wan2.1-1.3B and Cosmos-Predict2.5-2B with a short post-training pass and no added deployment cost \cite{chen2026data}. Trajectory Distribution Matching (TDM) unifies trajectory and distribution matching: a four-step PixArt-$\alpha$ student beats the teacher on Human Preference Score (HPS), Aesthetic Score (AeS), and user preference, and a four-step CogVideoX-2B student scores 81.65 on VBench against the teacher's 80.91 \cite{luo2025learning,huang2024vbench}. SANA-Sprint couples continuous-time consistency distillation with adversarial training and runs one-step 1024$\times$1024 generation in about 0.1 s on an H100, roughly 10$\times$ faster than FLUX-schnell \cite{chen2025sana}.

Recent studies also differ in where they impose a distribution objective. Phased DMD applies score matching within timestep subintervals, whereas TMD matches a coarse transition with a lightweight head that can perform an internal rollout~\cite{Fan_2026_CVPR,Nie_2026_CVPR}. For TMD the backbone evaluations and transition-head calls need separate accounting. The eDMD preprint trains a shared auxiliary score model on final student outputs and uses it in the updates for all denoising steps~\cite{zhao2026edmd}. These designs put different weight on intermediate trajectory alignment and final-sample matching. APT provides a comparison that post-trains a pretrained video generator against real data for one-step generation, while DOLLAR adds a latent reward to consistency and score distillation~\cite{pmlr-v267-lin25m,ding2025dollar}.

\subsubsection{Streaming Distillation}
\label{sec:sampling-autoregressive}

Teacher-forced autoregressive video training conditions on ground-truth history, while inference conditions on generated history. This mismatch can accumulate over long rollouts. CausVid converts a bidirectional diffusion transformer to a causal generator and applies four-step distribution matching with a KV cache~\cite{peebles2023scalablediffusionmodelstransformers,yin2025slow}. For 120-frame, $640\times352$ clips, its Table~3 reports 1.3\,s initial latency and 9.4 FPS throughput on an H100. These two measurements describe response and throughput, respectively; a whole-clip speedup requires the same end-to-end timing boundary for both models.

Self Forcing performs rollouts on student-generated history to close the training--inference gap: at 832$\times$480 it reaches 17.0 FPS in chunk mode with 0.69 s first-chunk latency, and 8.9 FPS frame-by-frame with 0.45 s per frame \cite{huang2026self}. Rolling Forcing adds a rolling window to suppress drift, reducing the long-horizon drift metric from 1.66 to 0.01 in a 30-second 832$\times$480 evaluation at 15.79 FPS and 0.76 s per chunk \cite{liu2026rolling}. Causal Forcing identifies a dependency mismatch when future-aware bidirectional teacher trajectories supervise a causal student; with a causal teacher plus DMD, Wan2.1-1.3B reaches 17.0 FPS and a VBench Quality score of 84.59 \cite{zhu2026causal}, and Causal-rCM pairs continuous-time consistency with self-forcing DMD \cite{zheng2026causal}. In streaming systems the gain comes jointly from causal structure, the KV cache, and the few-step student, so attributing FPS gains to distillation alone overstates those gains; first-chunk latency, steady-state rate, and full-generation time should be reported separately. Section~\ref{sec:systems} discusses the execution and scheduling costs behind those measures.

Reward Forcing uses motion-related rewards in distribution matching and updates a long-context attention sink~\cite{Lu_2026_CVPR}. The Self-Aligned Forcing preprint instead trains on differentiable history at the current denoising stage, aligning the noise level of the history with the active block~\cite{wang2026selfaligned}. These studies address related failures through different controls: the training objective for motion, the treatment of history, and the number of updates per generated chunk. A comparison needs clip length, resolution, denoising budget, and the timing boundaries for initial response, steady-state output, and the complete clip.

\subsection{Trajectory Reconstruction}
\label{sec:sampling-trajectory}

Trajectory reconstruction changes the path or how compute is allocated along it. The methods below reshape the transport path, learn transitions spanning time intervals, or switch resolution, denoiser, or representation across stages.

\subsubsection{Paths and Flow Maps}
\label{sec:sampling-geometry}

Trajectory geometry offers a complementary strategy: rather than fit a student to a curved multi-step path, change the path so that few-step integration is easier. Rectified Flow straightens trajectories through reflow \cite{liu2022flow}, and Flow Matching provides a general framework based on conditional probability paths and vector-field regression \cite{lipman2023flowmatchinggenerativemodeling}. InstaFlow straightens and then distills to one step: a one-step model reaches FID-5k about 23.3 and runs in 0.09 s on an A100, against 0.88 s and FID-5k 22.8 for a 25-step Stable Diffusion 1.4 (SD 1.4) baseline, where FID-5k denotes FID computed on 5k samples \cite{liu2023instaflow}. Straightening lowers curvature but does not by itself guarantee a lossless one-step map.

Interval mapping instead learns displacements spanning several steps directly. Shortcut Models condition on the target step size, so one training run serves arbitrary inference budgets \cite{frans2025one}; BOOT learns a time-conditioned student that reproduces the teacher output at any time step, bootstrapping its targets from two consecutive sampled steps so that no real data are needed \cite{gu2023boot}; MeanFlow regresses an interval-average velocity rather than an instantaneous one and reaches FID 3.43 at 1 NFE on ImageNet 256$\times$256 \cite{geng2026mean}. The three differ in conditioning variable and regression target: step size for Shortcut, target time (negative half log-SNR) for BOOT, and interval-average velocity for MeanFlow.

PTD parameterizes teacher trajectory segments with polynomials and queries teacher velocities along the path predicted by the student itself, keeping a curvature head only during training. On Wan2.1-14B its four-step model wins 55.1\% of blind human preferences against parallel decoding distillation, and on a LoRA of the 33B audio-video model MiniMax-H3 it wins 63.4\% against LightX2V Turbo, at the cost of teacher queries and polynomial fitting during training \cite{feng2026parametric}. Residual-optimization methods instead reshape error propagation for a given discretization, as ShortDF does by casting DDIM residual propagation as a shortest path on a reverse graph with edges $k<t$ \cite{chen2025optimizing}. These constraints reduce error at fixed NFE but do not guarantee that the output distribution is preserved. A supervision signal distorted by strong CFG can likewise propagate to the student.

Align Your Flow learns continuous-time maps between arbitrary time pairs, while FreeFlow distills a flow map from a pretrained model without access to real training data~\cite{sabour2025alignflow,Tong_2026_CVPR}. The AnyFlow preprint extends this to video and forms targets on student rollouts using shortcut flow-map transitions~\cite{gu2026anyflow}. Mean Flow Distillation uses average-velocity supervision to adapt pretrained flow-matching models to few-step generation~\cite{pmlr-v306-zhao26bx}. Compared with an endpoint-only mapping, these methods give the sampler a choice of interval lengths. Their evaluations should specify the test-time budget and the supervision available during training.

\subsubsection{Resolution Staging}
\label{sec:sampling-resolution}

Model-level cascading is the simplest multi-resolution scheme. Cascaded Diffusion and Imagen use a low-resolution model for global structure and a super-resolution model for detail; the former reports FID 1.48/3.52/4.88 on ImageNet 64/128/256, and the latter extends the cascade to 1024$\times$1024 \cite{ho2022cascaded,saharia2022photorealistic}. Most denoising then runs at low resolution, but each added stage brings another model and another source of cross-stage error.

A single model's trajectory can also be segmented by resolution. Pyramidal Flow treats the trajectory as a pyramid in which only the final stage is full-resolution, linking stages through probability-path interpolation and re-noising at jump points \cite{jin2025pyramidal}. Bottleneck Sampling alternates high, low, and high resolution, cutting the runtime of 50-step FLUX.1-dev from 33.85 s to 14.46 s \cite{tian2025training}. FlashVideo runs a low-resolution fidelity stage followed by a high-resolution detail stage, constructing a near-straight ODE from low- to high-resolution latents by flow matching; at 1080p it finishes in 102.3 s versus 2150 s for a single-stage 5B model \cite{zhang2026flashvideo}.

RMD aligns denoising intervals across resolutions by log-SNR and matches their distributions. It reports speedups of up to $33.4\times$ on SDXL and $25.6\times$ on Wan2.1-T2V-14B under the baselines in its Tables~1 and~2~\cite{chen2026cross}. The SelfLift preprint studies the transition itself: SelfLift-zero applies a training-free artifact correction, while SelfLift-rich trains a lightweight latent lifter and adapts on states visited by the student~\cite{wen2026selflift}. These methods address both the cost of high-resolution evaluations and the quality of the upsampled state. The achievable savings depend on resolution, schedule, and transition overhead.

\subsubsection{Model and State Switching}
\label{sec:sampling-representation}

T-Stitch reuses existing small and large pretrained denoisers on different stages of the same trajectory~\cite{pan2025tstitch}. SRDiffusion assigns video sketching to the large model and late rendering to the small model~\cite{cheng2025srdiffusion}. The TRACK preprint calibrates large--small disagreement and allocates the large model to individual steps, allowing non-contiguous schedules~\cite{munir2026track}. The TRACK evaluation uses the original sampler and step count; its reported $1.95$--$2.73\times$ speedup compares with an all-large video baseline in the authors' settings. The savings depend on model compatibility and the cost of switching. Structural reduction of an individual denoiser belongs to Section~\ref{sec:model-compression}.

In 3D generation, DreamFusion optimizes an asset per scene with score distillation sampling~\cite{poole2022dreamfusion}. DreamGaussian uses progressive densification of 3D Gaussians, while Instant3D generates sparse views and reconstructs a neural radiance field~\cite{tang2024dreamgaussian,kerbl20233d,li2024instant3d,mildenhall2021nerf}. These systems establish task-specific comparison points. Their full pipelines include different reconstruction or representation operations, so their wall times require explicit measurement boundaries.

TRELLIS uses two flow models for sparse structure and structured 3D latents~\cite{xiang2025structured}. MDT-dist distills both stages. In its Table~1, the TRELLIS teacher uses $25+25$ sampling steps and takes 6.1\,s on an A800. The $1+1$-step student takes 0.68\,s and the $2+2$-step student takes 0.94\,s, with different appearance and geometry scores~\cite{zhou2026few}. The table reports stage-wise sampling steps and latency. Converting steps to denoiser evaluations requires an explicit counting convention.

\subsection{Summary}
\label{sec:sampling-summary}

The three routes act on different terms of the cost model. Efficient sampling reduces $S$ or $n_k$ by changing the solver, schedule, or budget while reusing the backbone; few-step generation reduces $S$ to a few steps by changing the training objective; trajectory reconstruction can reduce the number of evaluations or lower their individual cost $C_{k,j}$ by reallocating work along the path. They therefore compose rather than sequence: a system may cut the teacher with a better solver, distill a few-step student, and then reduce stage overhead through cascading or caching.

Error sources differ accordingly. Efficient sampling is primarily affected by discretization error, vector-field curvature, and CFG strength; few-step generation additionally involves student capacity, supervision-target mismatch, teacher bias, and limited distribution coverage; staged generation must also handle path approximation, cross-resolution mismatch, and inter-stage error accumulation, with exposure bias and long-horizon drift in autoregressive video and multi-view consistency in 3D. No single FID or FPS value captures quality preservation; metrics and diagnostics should follow the task.

Efficiency metrics are meaningful only within their measurement boundaries. NFE counts function evaluations, not nominal steps, FLOPs, or latency; FPS may include caching, parallelism, and scheduling; end-to-end seconds must state whether encoding, decoding, reconstruction, and post-processing are included. Cross-paper comparisons should record model and data configuration, resolution or duration, CFG, the NFE counting rule, hardware, batch size, timing scope, and quality metric.

A practical selection order is to start with solver and schedule optimization when the backbone cannot be retrained; move to few-step distillation or native few-step training for single-digit budgets or interactive latency; and consider staged generation, specialization, caching, and system co-design when resolution, representation, or long-horizon structure dominates. Combined systems should locate the end-to-end bottleneck first and verify each module's marginal gain by ablation, rather than multiplying per-paper speedups.

\section{Efficient Attention and Sparse Computation}
\label{sec:attention}

Increasing spatial resolution or video duration increases the number of latent tokens processed at every denoising evaluation. For $N$ query and key tokens with head dimension $d$, dense self-attention takes $\mathcal{O}(N^2 d)$ arithmetic, and materializing the score matrix takes $\mathcal{O}(N^2)$ memory. Exact IO-aware implementations such as FlashAttention avoid materializing this matrix while evaluating the same dense interactions~\cite{dao2022flashattention}. This section studies algorithms that change those interactions or their computation in continuous image and video diffusion, including causal video diffusion~\cite{shao2026efficientvideodiffusionmodels}. Section~\ref{sec:systems} covers execution of exact and sparse operators, and Section~\ref{sec:dllm} covers diffusion language models.

\begin{figure}[!htbp]
    \centering
    \includegraphics[width=\linewidth]{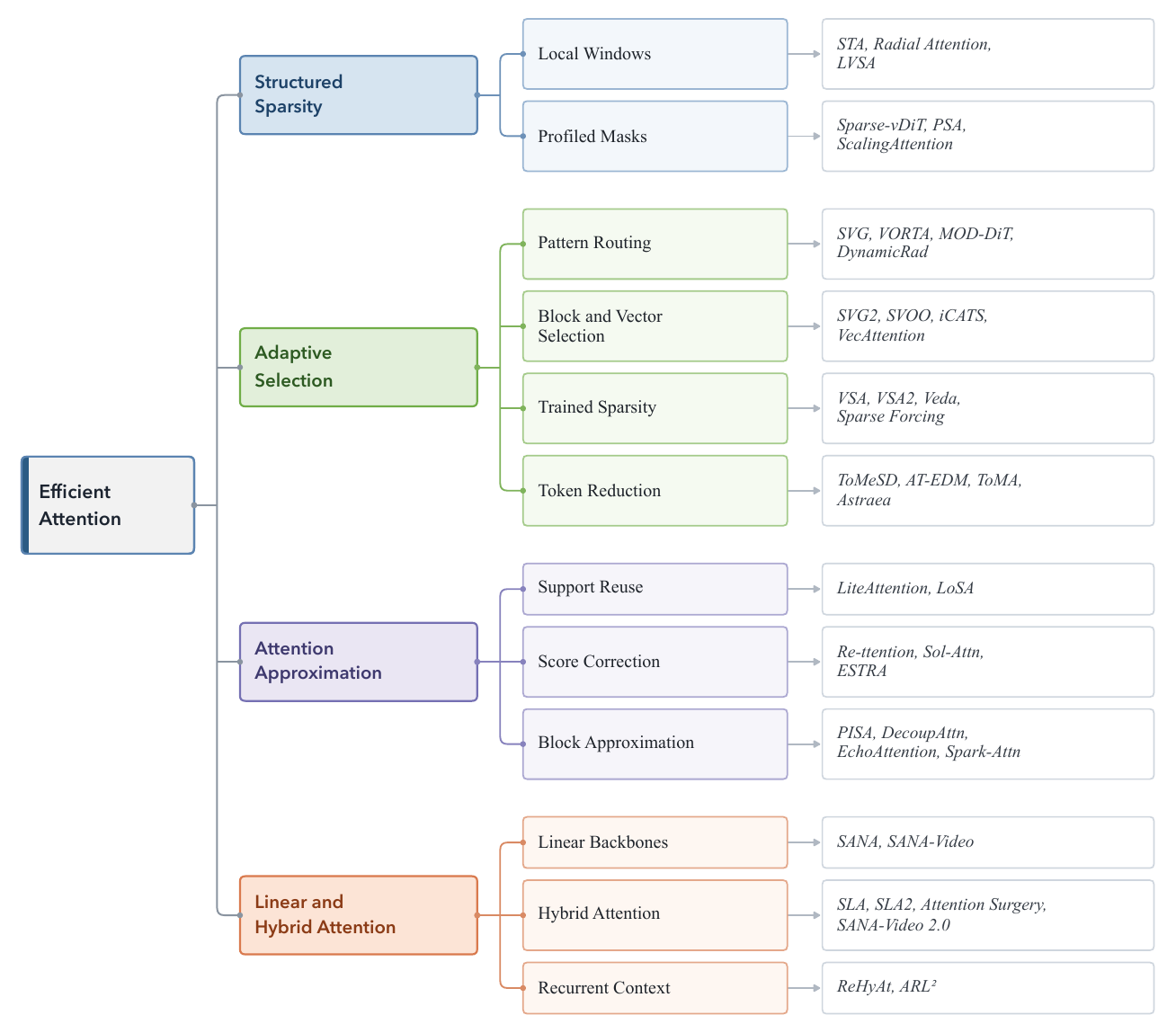}
    \caption{Taxonomy of efficient attention in image and video diffusion models. Leaf nodes show examples by primary mechanism.\ifsurveyappendices\space See Table~\ref{tab:attention-methods} for the full list.\fi}
    \label{fig:attention-taxonomy}
\end{figure}

\begin{figure}[!htbp]
    \centering
    \includegraphics[width=\linewidth]{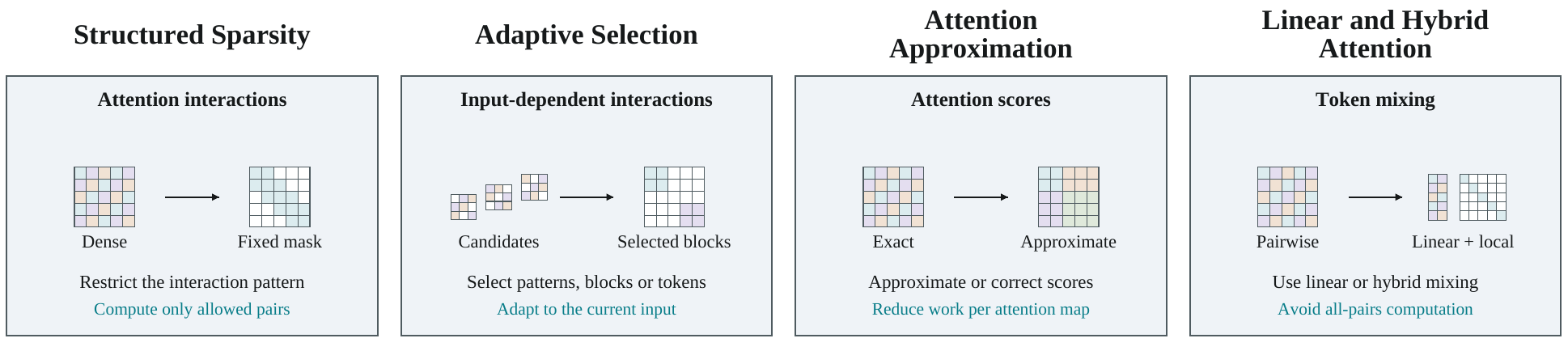}
    \caption{Mechanisms for efficient attention in diffusion models. Structured sparsity fixes the allowed interactions. Adaptive selection chooses interactions or tokens using input or inference state. Attention approximation estimates scores or reuses support. Linear and hybrid attention change the token-mixing operator.}
    \label{fig:attention-overview}
\end{figure}

\ifsurveyappendices Table~\ref{tab:attention-methods} in Appendix~\ref{app:methods:attention} compares each method's mechanism, adaptation requirements, and evaluation setting. \fi{}Figure~\ref{fig:attention-taxonomy} groups the methods by their primary intervention. Structured sparsity prescribes which interactions are available. Adaptive selection uses the current input, layer, or denoising stage to choose which interactions or tokens to compute. Attention approximation exploits stable selection patterns or estimates contributions that would otherwise be dropped. Linear and hybrid attention change the token-mixing operator. A model can use several of these mechanisms together. Figure~\ref{fig:attention-overview} uses attention maps and token-mixing operators to show where each mechanism reduces computation.

\subsection{Structured Sparsity}
\label{sec:attention-structured}

\subsubsection{Local Windows}
Video attention is often concentrated near the query in space and time. Sliding Tile Attention (STA) organizes a spatiotemporal window into contiguous tiles, so the kernel can execute regular matrix products for the retained interactions~\cite{pmlr-v267-zhang25m}. Radial Attention uses the empirical decay in attention weight with spatiotemporal distance to allocate a progressively smaller spatial neighborhood to more distant frames. Its long-video extension setting also uses fine-tuning~\cite{li2025radialattention}. LVSA uses local blocks and rotating globally visible anchor frames, giving the model a controlled path to distant context when a local window alone is insufficient~\cite{glorian2026lvsa}. These methods differ in how they allocate their long-range budget. Relevant quality measures include motion, subject identity, and consistency over the intended video duration.

\subsubsection{Profiled Masks}
Head-specific layouts address attention patterns that a single window cannot express. Compact Attention searches structured window parameters for individual heads and reuses the profiled configurations over denoising steps~\cite{https://doi.org/10.48550/arxiv.2508.12969}. Sparse-vDiT profiles diagonal, multi-diagonal, and vertical-stripe patterns and dispatches pattern-specific kernels~\cite{chen2026sparsevdit}. PSA parameterizes periodic stripes through widths, offsets, and strides, making several layouts executable with one kernel~\cite{jia2026parameterized}. ScalingAttention derives a sparse block topology from the weights and uses head-specific sensitivity to set its retained density~\cite{zhou2026scalingattention}. The amortized cost of profiling depends on how well a mask transfers to new prompts, resolutions, and sequence lengths.

In image generation, DiTFastAttn combines local windows and residual sharing with reuse across timesteps and guidance branches~\cite{yuan2024ditfastattn}. DiTFastAttnV2 uses head-wise arrow patterns and cache policies for multimodal diffusion attention~\cite{zhang2025ditfastattnv2}. These multi-component methods illustrate why a result for the whole configuration cannot isolate the effect of spatial sparsity. Section~6 discusses reuse of intermediate attention outputs and residuals.

\subsection{Adaptive Selection}
\label{sec:attention-adaptive}

\subsubsection{Pattern Routing}
Pattern routers choose from a small set of executable layouts. Sparse VideoGen (SVG) samples attention at inference time to decide whether a head should use a spatial or temporal pattern~\cite{pmlr-v267-xi25c}. VORTA routes among sparse variants designed to retain long-range video dependencies~\cite{sun2025vorta}. MOD-DiT models the attention pattern as a mixture, estimating the evolution from early denoising steps to reduce later profiling~\cite{pmlr-v306-liu26bb}. DynamicRad conditions a radial prior on semantic content and motion, with optional mask-aware adaptation~\cite{long2026dynamicrad}. Routing overhead is lowest when a decision applies to many interactions or remains valid for several steps. Rapidly changing attention and rare dependencies reduce the useful lifetime of such decisions.

\subsubsection{Block and Vector Selection}
Block selection constructs the sparse support for the current input. SVG2 uses semantic clustering and token permutation to arrange likely interactions into contiguous blocks, then applies a dynamic top-$p$ budget~\cite{yang2025sparsevideogen2}. SVOO separates offline layer-wise sensitivity from online query--key co-clustering~\cite{pmlr-v306-luo26a}. iCATS scores query--key interactions and varies the budget with the denoising timestep, while merging undersized clusters to limit padding~\cite{han2026icats}. HyperVAttention restricts clustering locally and updates assignments between anchor steps, trading refresh accuracy against selection cost~\cite{lee2026hypervattention}.

The sparse unit also determines execution cost. RainFusion2.0 uses block means and spatiotemporal reordering to build hardware-friendly masks~\cite{chen2025rainfusion20temporalspatialawarenesshardwareefficient}. DFSAttn uses a Hilbert ordering and hierarchical scoring to expose finer blocks~\cite{pmlr-v306-hu26p}. VecAttention selects informative vertical vectors and uses a matching sparse kernel~\cite{liu2026vecattention}. SpargeAttention screens blocks with a coarse estimate and a second online-softmax-aware filter~\cite{pmlr-v267-zhang25ch}. PAROAttention co-designs pattern reordering and low-bit attention~\cite{zhao2025paroattention}. Reported density is useful alongside indexing, permutation, padding, and gather costs, since a finer mask can save arithmetic while increasing data movement. Section~\ref{sec:model-compression} treats the precision axis of such combinations.

\subsubsection{Trained Sparsity}
Training with sparse computation allows the model to adapt to its support. VSA routes at the tile level and computes token-level attention within selected blocks~\cite{zhang2025fastervideodiffusion}. VSA2 introduces finer routing, variable budgets, and sparse rebasing during progressive training~\cite{zhang2026vsa2}. BSA reduces work on both sides of attention by selecting queries and key--value blocks~\cite{zhan2026bsa}. VMoBA selects blocks with layer-wise partitions, whereas MoGA learns token groups for semantically related interactions~\cite{wu2026vmoba,jia2026moga}. Veda distills tile-level scoring and head-aware patterns from dense attention~\cite{pmlr-v306-han26y}. For these methods, training FLOPs, realized training time, and deployment latency answer separate questions.

Causal video generation introduces a growing history and an asymmetric interaction pattern. Light Forcing allocates sparsity by chunk and selects frames and blocks hierarchically~\cite{pmlr-v306-lv26i}. Sparse Forcing learns to retain persistent salient blocks and routes locally within the current window~\cite{xu2026sparseforcing}. DASA combines motion-driven regional computation with a policy for retaining historical KV blocks~\cite{he2026dasa}. Comparing these methods requires the same context horizon and decoding configuration. BLADE jointly trains block sparsity and step distillation~\cite{gu2026blade}. Its joint acceleration combines cheaper denoiser evaluations with a reduced evaluation count, the complementary mechanism in Section~\ref{sec:sampling}.

\subsubsection{Token Reduction}
Token reduction shrinks the sequence passed through attention or surrounding blocks. ToMeSD merges similar spatial tokens and restores the layout after computation~\cite{bolya2023tomesd}. AT-EDM uses attention-derived saliency to prune redundant tokens during inference~\cite{wang2024atedm}. Structure-then-Detail distinguishes structural content from local detail when deciding which tokens can be merged~\cite{fang2025structuredetail}. ToMA selects diverse representatives and casts merging and restoration as matrix operations, reducing the overhead of a plug-in token reducer~\cite{pmlr-v267-lu25v}. Astraea extends input-dependent token selection to video DiTs with timestep-specific budgets and parallel sparse execution~\cite{liu2026astraea}. The deployment gain depends on the total cost of selection, reduced computation, and restoration. Persistent changes to the latent representation are covered in Section~\ref{sec:model-compression}.

\subsection{Attention Approximation}
\label{sec:attention-temporal}

\subsubsection{Support Reuse}
LiteAttention identifies tiles that can be skipped early in denoising and carries the skip decisions into later steps~\cite{shmilovich2025liteattention}. LoSA constructs a block support at a dense step using a retained-attention-mass criterion, then reuses the selected indices~\cite{liu2026losa}. Both methods still compute attention on the selected interactions at the current step. Their savings come from reducing both the selected work and the cost of discovering the mask again. The relevant failure mode is a stale support that omits interactions needed as an object's appearance or motion develops. Section~6 treats reuse of the computed attention output or historical KV tensors.

\subsubsection{Score Correction}
Masking changes the softmax normalization as well as the evaluated dot products. For a query $q$, let $s_j=q^\top k_j/\sqrt{d}$ and let $\mathcal{S}$ be the exact support. Then the dense output is
\[
  y(q)=
  \frac{\sum_{j\in\mathcal{S}} e^{s_j}v_j + U_{\bar{\mathcal{S}}}}
       {\sum_{j\in\mathcal{S}} e^{s_j} + Z_{\bar{\mathcal{S}}}},
  \quad
  U_{\bar{\mathcal{S}}}=\sum_{j\notin\mathcal{S}}e^{s_j}v_j,\quad
  Z_{\bar{\mathcal{S}}}=\sum_{j\notin\mathcal{S}}e^{s_j}.
\]
An approximation can reduce the error from replacing the complement by zero by estimating its normalization or weighted value. Re-ttention uses historical softmax statistics to reshape the current sparse distribution~\cite{chen2025rettention}. Sol-Attn filters blocks during online softmax and uses proxy scores to account for omitted blocks~\cite{li2026solattn}. ESTRA uses query representatives to score keys and applies a variance-based correction to the summarized contributions~\cite{liu2026estra}. Attention mass is a useful local criterion, but its relationship to visible output changes is empirical. Causal evidence on attention sinks motivates validating quality at the final output as well as at the operator~\cite{pmlr-v306-wu26ax}.

\subsubsection{Block Approximation}
PISA combines exact attention on selected blocks with a Taylor approximation for the other blocks, using a shared online normalization~\cite{li2026pisa}. EchoAttention exploits similarity between frame-block attention distributions and learns routing through distillation~\cite{pmlr-v306-xia26c}. DecoupAttn uses semantic summaries over all keys and applies exact corrections to selected spatial blocks, decoupling the partition for summaries from the exact support~\cite{li2026decoupattn}. Spark-Attn uses attention preferences to regroup tokens into regular blocks and reweights the compressed summaries~\cite{tang2026sparkh3}. These approaches differ in how they amortize the cost of approximating the rest of the sequence. Compare against hard masking at a matched exact density, account for the compute spent on summaries, and report both fidelity to dense generation and task-level quality.

\subsection{Linear and Hybrid Attention}
\label{sec:attention-linear}

\subsubsection{Linear Backbones}
With a feature map $\phi$ and kernel $\kappa(q,k)=\phi(q)^\top\phi(k)$, normalized kernel attention can accumulate sufficient statistics before applying the queries. The numerator uses $\phi(q)^\top\sum_j\phi(k_j)v_j^\top$, and the denominator uses $\phi(q)^\top\sum_j\phi(k_j)$. For fixed feature and head dimensions, this calculation scales linearly in sequence length. The choice of feature map, normalization, and training matters because the operator changes the learned token interactions. SANA uses a linear DiT to scale text-to-image generation to high resolutions, alongside latent compression and changes to the sampling pipeline~\cite{xie2025sana}. SANA-Video uses block-linear attention and maintains recurrent statistics for causal video generation~\cite{chen2026sanavideo}. Across such architectures, matched-backbone ablations help separate attention gains from changes to the autoencoder, model size, and sampling steps.

\subsubsection{Hybrid Attention}
Hybrid methods reserve exact softmax for selected interactions or layers. SLA computes important interactions with sparse softmax and uses a linear path for weaker interactions~\cite{zhang2026sla}. SLA2 adds a learned router, learnable mixing, and quantization-aware fine-tuning~\cite{zhang2026sla2}. Attention Surgery allocates softmax and linear computation by token block and adapts a pretrained video DiT through lightweight distillation~\cite{ghafoorian2026attentionsurgery}. SANA-Video 2.0 trains a hybrid backbone with periodic softmax layers and attention residuals across depth~\cite{chen2026sanavideo2}. The first three methods allocate work within layers; the last uses placement across layers to retain expressivity. Their efficiency depends on whether the expensive path can stay sparse and whether the linear path preserves global composition and temporal structure.

\subsubsection{Recurrent Context}
ReHyAt combines local softmax with global linear mixing in a chunked recurrent form, adapting a pretrained model through distillation~\cite{ghafoorian2026rehyat}. ARL$^2$ keeps softmax for intra-frame interactions, summarizes prior frames in a fixed-size gated linear state, and updates that state after denoising the frame~\cite{li2026attend}. A fixed-size state controls the memory growth of the linear branch. Its ability to preserve fine historical detail should be evaluated over the intended horizon, with separate reporting for state size, chunk latency, and video consistency.

\subsection{Summary}
\label{sec:attention-summary}

The four routes trade computation for different assumptions about the structure of video and image attention. Structured masks amortize selection; dynamic masks follow the input at a routing cost; complement approximations spend extra work to reduce the error of hard masking; and linear or hybrid operators change how the model represents interactions. Attention density alone cannot determine the benefit of any route. For inference comparisons, report the backbone, resolution, frame count, denoising steps, precision, GPU, and dense baseline, and distinguish attention-kernel, denoiser, and full-pipeline latency. For trained methods, also report adaptation cost and changes to the deployed model. In long-video settings, temporal coherence and fidelity over the intended horizon can reveal failures missed by short-clip averages.

Execution can still limit gains at the same density. Distributed Sparse Attention (DSA) and FlashOmni address parallel execution and sparse-kernel dispatch, complementary to the interaction choices reviewed here~\cite{li2026dsa,qiao2025flashomniunifiedsparseattention}. Cross-axis results should identify which components were active so that reductions in attention work, sampling steps, and data movement can be interpreted separately.

\section{Caching and Computation Reuse}
\label{sec:caching}

Each denoising step supplies a new latent to the same network. Some intermediate states evolve slowly enough that a previous computation can replace part of a later forward pass~\cite{ma2023deepcache,wimbauer2023cache,liu2025cachesurvey}. A stream of frames or requests creates additional opportunities for reuse. The relevant question is which dependencies of a cached state may change before it must be recomputed. A denoising feature typically changes with the latent and timestep, a causal history can remain fixed as new frames arrive, and an activation retrieved from another request must satisfy a compatibility test.

\begin{figure}[!htbp]
    \centering
    \includegraphics[width=\linewidth]{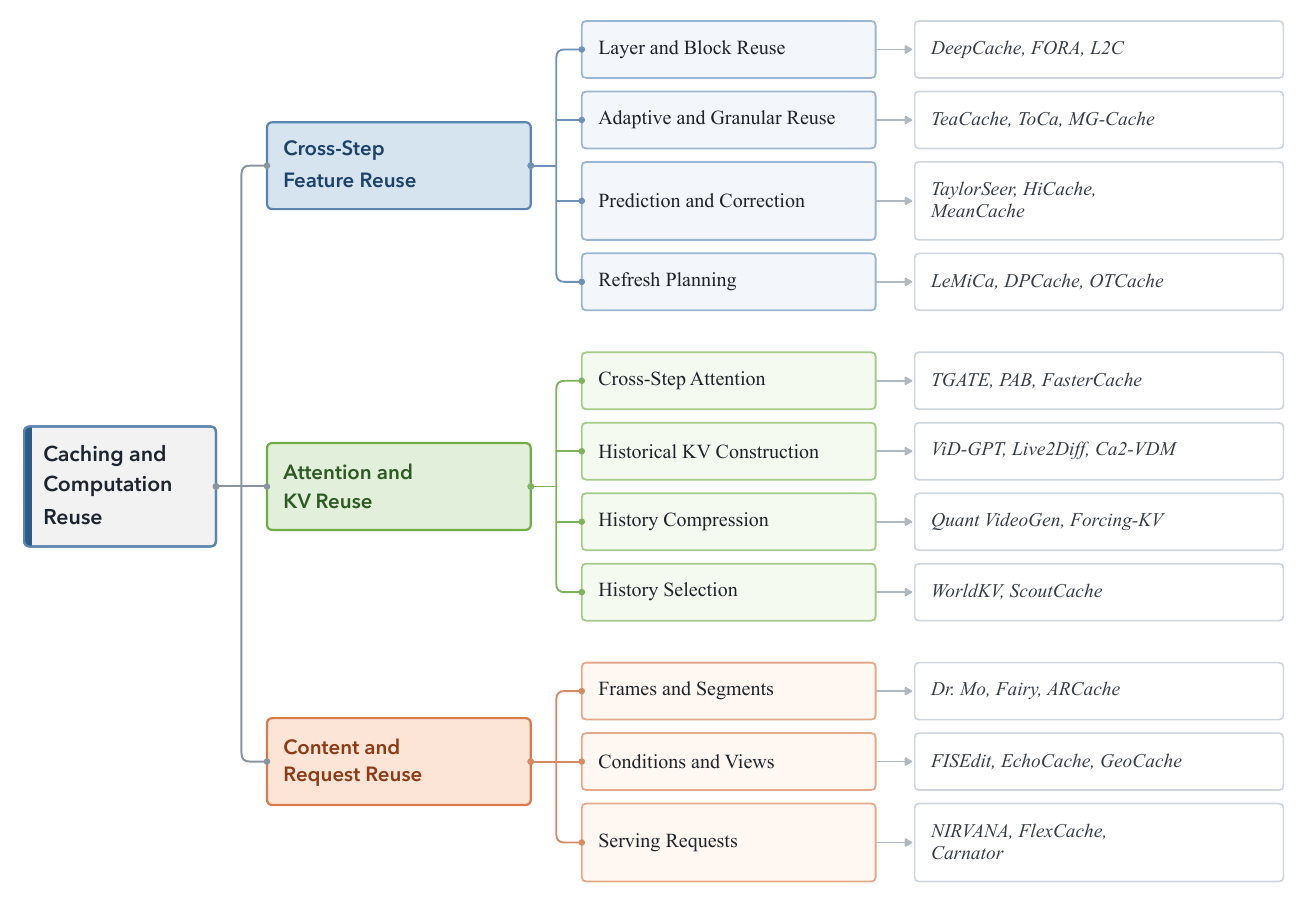}
  \caption{Taxonomy of caching and computation reuse. Each route branches into subtypes with representative methods listed at the leaves.\ifsurveyappendices\space Table~\ref{tab:caching:methods} provides the full list.\fi}
    \label{fig:caching:taxonomy}
\end{figure}

\begin{figure}[!htbp]
    \centering
    \includegraphics[width=\linewidth]{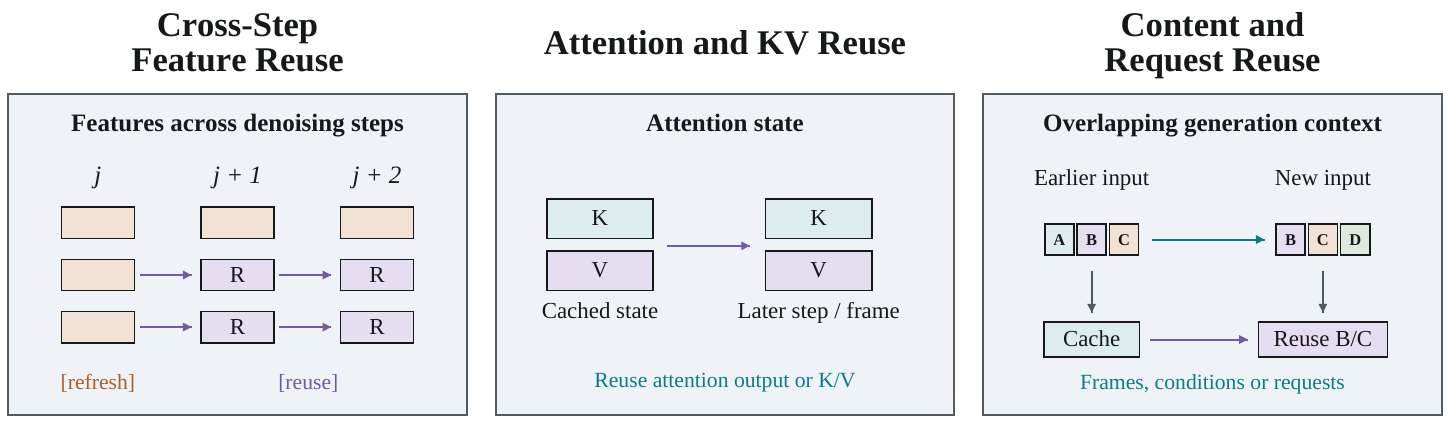}
    \caption{Three routes to computation reuse. Cross-step feature reuse refreshes or reuses intermediate features. Attention and KV reuse carries forward attention outputs or key--value state. Content and request reuse retrieves compatible state across frames, conditions, or requests. R marks a reused feature.}
    \label{fig:caching-overview}
\end{figure}

\ifsurveyappendices Table~\ref{tab:caching:methods} in Appendix~\ref{app:methods:caching} compares methods by cached state, refresh or selection signal, execution, and evaluation setting. \fi{}Figure~\ref{fig:caching:taxonomy} shows the three routes. Cross-step feature reuse reduces the cost of denoiser evaluations. Attention and KV reuse either bypasses attention submodules or avoids rebuilding historical context. Content and request reuse transfers state across frames, conditions, views, or requests. Composite methods are listed under their primary mechanism. Figure~\ref{fig:caching-overview} illustrates which states can be reused across denoising steps, across attention evaluations, and across overlapping inputs.


\subsection{Cross-Step Feature Reuse}
\label{sec:caching:features}

Let \(m_{t,l}=1\) if denoiser module \(l\) is executed at step \(t\), and let its cost under the current execution shape be \(c_{t,l}\). On a fixed sampling grid with one denoiser evaluation per step, a schematic compute model is
\begin{equation}
  C_{\mathrm{full}} = \sum_{t,l} c_{t,l},\qquad
  C_{\mathrm{reuse}} \approx \sum_{t,l}m_{t,l}c_{t,l}+C_{\mathrm{manage}},
  \label{eq:caching:cost}
\end{equation}
where \(C_{\mathrm{manage}}\) covers cache reads and writes, change estimation, prediction, and scheduling. The same accounting can index individual evaluations for samplers with multiple evaluations per step. A cache hit can bypass a block, token group, or most of a backbone invocation while the sampler still advances its latent on the prescribed grid. The full-evaluation count, nominal step count, and measured latency therefore describe different aspects of cost. Small token groups can save theoretical operations while gathering, scattering, and kernel launches erode the realized benefit.

\subsubsection{Layer and Block Reuse}

\textit{DeepCache} exploits the U-Net hierarchy. During reuse steps, it updates the shallow path and supplies the deeper semantic feature from an earlier step. In its Stable Diffusion v1.5 setting, the authors report a \(2.3\times\) acceleration with a small CLIP-score reduction~\cite{ma2023deepcache}. \textit{Faster Diffusion} uses the encoder--decoder split to reuse encoder features at selected steps. \textit{FRDiff} studies how feature reuse can complement a low-NFE sampler~\cite{li2024fasterdiffusionrethinkingrole,so2024frdifffeaturereuse}. The latter needs separate sampler-only and reuse-only baselines to quantify the contribution of each intervention.

A Transformer admits different cut points. \textit{FORA} caches attention and MLP outputs, while \textit{\(\Delta\)-DiT} reuses block increments so the residual stream can still carry the current input~\cite{selvaraju2024fora,chen2024delta}. \textit{Cache Me if You Can} derives block-level plans from measured feature dynamics, and \textit{L2C} learns lightweight routers for time-dependent layer selection~\cite{wimbauer2023cache,ma2024learningtocache}. These designs expose two transferable choices: the reusable state can be a whole output or a residual contribution, and the chosen cache locations can be fixed, profiled, or learned.

\subsubsection{Adaptive and Granular Reuse}

Online policies use a cheap signal to decide whether the next expensive computation is worth executing. \textit{TeaCache} estimates residual change from the timestep-modulated input and accumulates the estimate until a threshold triggers a refresh. \textit{MagCache} uses the magnitude evolution of residuals and calibration statistics. \textit{DiCache} probes shallow blocks to decide whether to reuse deeper computation~\cite{DBLP:journals/corr/abs-2411-19108,ma2025magcache,bu2025dicache}. \textit{AdaCache} adapts the refresh interval to feature change and video motion, while \textit{Foresight}, \textit{BWCache}, and \textit{LightCache} exploit layer variation, block stability, and stage-dependent cache policies, respectively~\cite{kahatapitiya2024adaptive,adnan2025foresight,cui2025bwcacheacceleratingvideodiffusion,https://doi.org/10.48550/arxiv.2510.05367}. The proxy should be cheaper than the skipped computation and remain calibrated as the checkpoint, schedule, or clip dynamics change.

Fine-grained policies use the fact that different units in the same step can tolerate different reuse intervals. \textit{ToCa} ranks tokens by sensitivity, and \textit{DuCa} alternates coarse and fine reuse~\cite{zou2024accelerating,zou2024DuCa}. \textit{RAS} updates selected regions and reuses noise predictions elsewhere, placing it at the intersection of reuse and the selective computation in Section~\ref{sec:attention}. \textit{BlockDance} uses spatiotemporal similarity, and \textit{ProfilingDiT} exploits the different dynamics of foreground and background video features~\cite{ras2026,zhang2025blockdancereusestructurallysimilar,ma2025model}. Task structure also informs which units to refresh. \textit{Fast3Dcache} protects changing geometry boundaries, \textit{MoECa} makes reuse decisions at expert granularity, and \textit{WorldCache} accounts for heterogeneous tokens within a world-model denoising step~\cite{yang2026fast3dcachetrainingfree3dgeometry,li2026moecaaligningfeaturereuse,feng2026worldcacheacceleratingworldmodels}.

\textit{MG-Cache} combines offline timestep profiling with adaptive layer-wise selection of attention and MLP submodules~\cite{chen2026mgcache}. It illustrates why step-level and layer-level decisions can benefit from different signals. \textit{DiSC} couples token and sparsity-mask reuse with a hardware dataflow so the selected work can be skipped by the accelerator~\cite{disc2026}. Both examples highlight the need to measure how the execution backend handles the chosen granularity.

Thresholding and selective computation can be compared along two independent dimensions: how often the cache is refreshed and how much work remains at a refresh. A model-level threshold can leave each executed forward pass on a regular kernel path and therefore has low control overhead. A token-level policy can preserve changing regions under the same operation budget, but only if the implementation turns the mask into smaller executable work. In DiTs, updating a subset of queries also requires a decision about whether its keys and values come from current or cached states. That choice affects both the cost model and the fidelity of interactions between updated and reused tokens. For video, comparing policies over clips with different motion levels is more informative than setting a single global refresh interval and averaging the results.

\subsubsection{Prediction and Correction}

Simple reuse supplies the most recent cached feature. Its local error grows when feature derivatives are large or refreshes are far apart. To see how this can affect the whole trajectory, consider one explicit sampling update \(x_{t+1}=x_t+h_tv_\theta(x_t,t)\). Let \(\widetilde{x}_t\) denote the cached trajectory and \(\epsilon_t\) the cache error evaluated at that trajectory. If \(v_\theta\) is locally \(L_t\)-Lipschitz over the two states, then the state deviation \(\delta_t=\|\widetilde{x}_t-x_t\|\) satisfies
\begin{equation}
  \delta_{t+1}
  \leq (1+|h_t|L_t)\delta_t+|h_t|\|\epsilon_t\|.
  \label{eq:caching:drift}
\end{equation}
This illustrative bound applies to the stated update. It explains why the same feature error may have different downstream effects at different points on the trajectory. A full refresh removes the approximation from the current evaluation and still receives the latent produced by earlier approximate steps.

\textit{TaylorSeer} estimates temporal feature derivatives from refresh history and extrapolates via a Taylor expansion. \textit{AB-Cache} combines historical states with an Adams--Bashforth predictor, and \textit{EOC} estimates and compensates the error of reused features~\cite{Liu_2025_ICCV,yu2025ab,qiu2025acceleratingdiffusiontransformererroroptimized}. \textit{RFC} uses module input--output relations to predict reuse error, while \textit{ERTACache} couples error rectification with adjustments to the timestep schedule~\cite{son2026relationalfeaturecachingaccelerating,peng2025ertacache}. These approaches spend additional computation or memory on approximating the current state more closely.

Recent methods modify how forecasting is parameterized and corrected. \textit{FoCa} treats the hidden-state sequence as a feature ODE and calibrates the forecast. \textit{HiCache} uses scaled Hermite polynomials as its forecasting basis. \textit{SVD-Cache} forecasts dominant feature subspaces and uses a separate policy for the residual. \textit{Z-Cache} uses a subsequent full-computation anchor to assess and correct deviations from a provisional cache window~\cite{zheng2026foca,feng2026hicache,Chen_2026_CVPR,cheng2026zcache}. This makes forecast error, correction cost, and the timing of the next anchor distinct design variables.

Flow-matching methods can also approximate the quantity used directly by the solver. \textit{MeanCache} reuses Jacobian--vector products to estimate interval-average velocity and schedules refreshes using trajectory stability~\cite{meancache2026}. \textit{TACache} constructs an offline schedule using magnitude and direction variation, then reconstructs skipped velocities using the sample's recent history~\cite{liu2026tacache}. The former changes the quantity being approximated, while the latter decomposes the error of a point-velocity update. The trajectory and sampler affect both approaches. With stochastic sampling, the noise realization also influences subsequent states, so transfer from a deterministic sampler requires direct evaluation.

A matched comparison illustrates the distinction between forecasting quality and runtime. In the \textit{SVD-Cache} paper's FLUX.1-dev experiment, the 50-step baseline took 25.82\,s. At the same reported 4.16-fold reduction in reported FLOP count, \textit{TaylorSeer} took 7.46\,s and produced an ImageReward of 0.9768, while \textit{SVD-Cache} took 7.62\,s and produced 1.0123; the baseline score was 0.9898~\cite{Chen_2026_CVPR}. This operating point shows why forecast policies should be compared with both measured latency and a quality metric. A similarly useful comparison for video measures motion and temporal consistency, since frame-wise similarity can miss changes to motion introduced by forecast drift.

\subsubsection{Refresh Planning}

A local threshold reacts to the current request. Offline planners allocate a budget across the full trajectory. \textit{LeMiCa} estimates terminal error for candidate reuse intervals, represents them as edges in a directed acyclic graph, and minimizes the largest profiled edge errors lexicographically. It reports \(2.9\times\) acceleration on Latte~\cite{lemica2025}. \textit{DPCache} also casts refresh scheduling as denoising path planning. The trajectory-oriented \textit{OmniCache} studies global reuse patterns along the sampling path~\cite{cui2026denoisingpathplanningtrainingfree,chu2025omnicache}. \textit{OTCache} uses a conservative reference schedule and a searched low-budget anchor to interpolate schedules for intermediate budgets~\cite{otcache2026}. The benefit of these policies depends on how well a profiled or searched schedule transfers to held-out prompts and other sampler settings. Minimizing a profiled edge cost controls the planning objective and offers empirical evidence about final output fidelity.

Other methods broaden what the scheduler allocates. \textit{InvarDiff} profiles step, layer, and module choices; \textit{SADA} couples step and token decisions using ODE stability; \textit{ReCache} learns budget-conditioned schedules; and \textit{EVO} searches block--timestep schedules for diffusion policies~\cite{wu2026finegrainedcachingdiffusiontransformers,sada2025,aliev2026recachelearningbudgetawarecaching,wang2026evolvingcacheschedulesfast}. \textit{BeCARE} brings error amplification, cache age, and budget feedback into refresh decisions~\cite{zhang2026becarebudgetedcacherefresh}. These methods differ in the calibration and search they require before serving, and in how much work remains on the critical path of each request.

Refresh planning also has an amortization boundary. A grid profiled once for a fixed checkpoint and sampler can serve many requests, whereas a policy conditioned on the current trajectory may need probes at inference time. Evaluating the offline search time, the online decision time, and the range of budgets served by a single plan makes this trade-off visible. It also reveals when layer-level and trajectory-level choices interact. For example, an aggressively compressed grid makes successive states farther apart, which can change the ranking of otherwise cacheable blocks. A useful comparison therefore reports how performance changes when the schedule is retuned and when an already calibrated schedule is transferred.

\subsection{Attention and KV Reuse}
\label{sec:caching:attention}

\subsubsection{Cross-Step Attention}

An attention-output cache bypasses selected attention computations by supplying a previous output or a forecast. Both the queries and the attention map can evolve as the latent changes. \textit{TGATE} uses different policies for cross-attention during semantic formation and self-attention during subsequent refinement~\cite{liu2024faster}. \textit{PAB} exploits distinct temporal stability in spatial, temporal, and cross-modal attention and sets separate broadcast intervals. Its strongest reported system results also use broadcast sequence parallelism~\cite{zhao2024real}.

\textit{FasterCache} combines dynamic attention-feature reuse with classifier-free guidance acceleration, and \textit{UniCP} combines caching with pruning of attention components~\cite{lv2024fastercache,sun2025unicpunifiedcachingpruning}. These components can interact with attention kernels and the two CFG branches. A cache-only ablation and a full-system measurement answer different questions about their contribution.

The scope of the cached state determines the usable kernels. Reusing an entire attention output can bypass attention and its surrounding projections. Preserving a subset of heads or tokens instead requires assembling reused and recomputed results. The decomposition matters most in video, where temporal and spatial attention can exhibit different stability and tensor sizes. Evaluations can show which operation was removed by reporting attention time and peak allocated memory in addition to the overall denoising time, with the same parallelism and attention implementation across baselines.

\subsubsection{Historical KV}

A causal model can retain keys and values of past frames or chunks once their states, positions, conditioning, and attention dependencies remain fixed under its execution schedule. For new queries at a particular layer, the dense attention operation remains
\begin{equation}
  A_{\mathrm{new}} =
  \operatorname{Attn}\!\left(
    Q_{\mathrm{new}},[K_{\mathrm{hist}},K_{\mathrm{new}}],
    [V_{\mathrm{hist}},V_{\mathrm{new}}]\right).
  \label{eq:caching:history}
\end{equation}
Retaining compatible historical KV avoids projecting it again. With full attention, the new queries still read and attend to the retained history. Cache capacity and memory traffic accordingly grow with context length. If past tokens are re-noised or their positional or conditioning inputs change, reuse requires matching the relevant state or introducing an approximation. Section~\ref{sec:dllm:acceleration} discusses the corresponding challenge when masked language diffusion uses bidirectional context that continues to change.

\textit{ViD-GPT}, \textit{Live2Diff}, and \textit{Ca\(^2\)-VDM} build historical reuse into causal, unidirectional, or cache-sharing video architectures~\cite{https://doi.org/10.48550/arxiv.2406.10981,xing2024live2difflivestreamtranslation,https://doi.org/10.48550/arxiv.2411.16375}. At longer horizons, \textit{Quant VideoGen} uses 2-bit semantic-aware KV quantization with progressive residuals, while \textit{Forcing-KV} adapts pruning to head behavior~\cite{xi2026quantvideogenautoregressivelong,ji2026forcingkvhybridkvcache}. Compression raises a quality question about whether the retained state preserves motion, subjects, and long-range dependencies.

\textit{FAST-AR} uses temporal correspondence to compress KV and combines this with sparse attention~\cite{samuel2026fastautoregressivevideodiffusion}. \textit{WorldKV} retrieves historical memory relevant to the current viewpoint or action~\cite{yi2026worldkvefficientworldmemory}. \textit{ScoutCache} addresses all three costs in few-step autoregressive video generation: current denoising, historical KV reads, and construction of new cache entries. It uses a shallow readout for selected denoising updates, budgets reads by attention-output risk, and selects representative spatial tokens for storage. Its preprint reports up to \(2.28\times\) and \(1.86\times\) speedups on Self-Forcing and LongLive, respectively, for the combined system~\cite{tan2026scoutcache}. The distinction between storage, read traffic, and attention work determines which intervention helps at each sequence length.

For a uniform dense cache with \(L\) cached layers, \(N\) retained tokens, \(H_{\mathrm{kv}}\) KV heads, head width \(d_h\), and \(b\) bytes per stored element, the tensor payload alone is \(2LNH_{\mathrm{kv}}d_hb\) bytes per sequence. Quantization mainly changes \(b\), retention and retrieval change \(N\), and head-specific compression can alter both. Auxiliary indices, quantization scales, padding, and transfer buffers contribute additional memory. At fixed \(N\), a representation with smaller storage may still be slower if decoding it adds work on the critical path. Conversely, transferring a small, relevant subset can reduce read traffic and attention work together when the backend operates directly on the subset.

\subsection{Content and Request Reuse}
\label{sec:caching:content}

\subsubsection{Across Frames and Segments}

\textit{VidRD} conditions clip extension on reused latent features from an existing clip and adapts the decoder for temporal continuity~\cite{vidrd2023}. \textit{Dr. Mo} propagates early coarse denoising states from anchor frames using motion estimates, then resumes regular denoising on the remaining steps~\cite{drmo2024}. \textit{Fairy} caches anchor-frame K/V for video editing and uses cross-frame attention to transfer the edit to other frames. Its paper reports 14 seconds for editing 120 frames of \(512{\times}384\) pixels in a system that also adapts the model and parallelizes edits~\cite{fairy2024}. The hierarchical \textit{OmniCache} combines frame, token, block, and layer-level reuse~\cite{omnicache2026}. Clip boundaries, disocclusions, and large motion test the correspondence that supports these methods.

\textit{ARCache} adapts a different kind of state across video segments. It uses previous-segment behavior to schedule reuse within the current segment and corrects the residual trajectory for later segments~\cite{Nan_2026_CVPR}. Previous segments become evidence for the refresh policy, in addition to serving as generation context. This matters because approximate denoising in one segment can affect both its own visual quality and the conditioning on which the next segment is generated. Evaluation over a sequence of segment boundaries can reveal progressive drift that a single-clip comparison misses.

\subsubsection{Across Conditions and Views}

Edit masks, external modalities, and camera geometry provide explicit correspondence signals. \textit{FISEdit} retains features outside the affected image regions during text-guided refinement~\cite{Yu2023FISEditAT}. \textit{EchoCache} uses audio energy to guide refreshes of video latents. \textit{CAVC} uses the convergence of the audio branch to suspend audio computation during middle steps, then restores joint denoising to recover audio--video alignment~\cite{chen2026echocacheenergyguidedcrossmodalcaching,cavc2026}. These approaches use different information: external saliency guides visual refresh in the first case, and modality-specific convergence guides branch scheduling in the second.

\textit{GeoCache} refreshes selected anchor views and transports clean-signal updates between corresponding surface points. Periodic full-view evaluation limits drift when correspondences become unreliable~\cite{li2026geocachetrainingfreeaccelerationmultiview}. Reported denoiser-loop gains describe one part of a pipeline that also constructs and renders textured assets. Quality checks should capture view consistency as well as per-view image fidelity.

\subsubsection{Across Serving Requests}

Cross-request reuse retrieves a previous generation state and uses it to skip work for a new request. \textit{NIRVANA} retrieves intermediate image diffusion latents from semantically similar prompts. Its production workloads show a 19.8\% end-to-end latency reduction and 21\% less GPU computation~\cite{agarwal2023approximatecachingefficientlyserving}. \textit{FlexCache} extends this idea to video through cache compression and object--background decoupling~\cite{flexcache2025}. \textit{Carnator} computes a signature from a lightweight early denoising probe to judge generation-level compatibility before reusing a latent trajectory and sparse attention connections. The reported \(2.17\times\) maximum is a cache-hit end-to-end speedup~\cite{carnator2026}. Workload latency additionally depends on hit rate, the cost of rejected probes, retrieval overhead, and the distribution of misses. Conditioning and output metrics should also capture whether reused content leaks into requests whose conditions differ.

The definition of a hit becomes part of the method in this setting. A more permissive threshold increases the number of reusable requests and can also increase mismatches. A stricter threshold can produce a faster subset of accepted requests while leaving much of the workload on the dense path. Evaluating both policies at a fixed quality target, with preceding requests replayed in chronological order and storage capacity held constant, reveals the trade-off. Cache warm-up and maintenance costs become material when popular requests or generation parameters change over time.

\subsection{Evaluation and Trade-Offs}
\label{sec:caching:evaluation}

For \textbf{cross-step reuse}, a useful comparison holds the checkpoint, input, sampler, guidance, resolution, and timing boundary constant, then sweeps the refresh budget. Report nominal steps, full and partial denoiser evaluations, end-to-end latency, peak memory, and quality. Pair baseline-referenced measures such as LPIPS with task metrics such as prompt alignment or temporal coherence. Comparing at similar latency exposes the quality cost of different approximations. Combining caching with fewer sampling steps warrants a matched ablation because shorter trajectories change the span and predictability of each skipped evaluation.

For \textbf{long video and world models}, report initial latency, steady-state throughput, memory versus generated duration, and consistency after long rollouts or revisits. A short VBench clip can capture local quality while failing to test the recall of a previously observed scene. Measure KV read and write costs alongside cache occupancy, especially when compression, sparse attention, and denoising shortcuts are used together. For \textbf{serving}, distinguish hit latency, miss latency, and full-workload throughput. Report the calibration or profiling cost, cache capacity, and hit rate with a stated workload and quality threshold. These settings favor different designs because they place the main cost in repeated forwards, historical context, or request-level retrieval.

Across these routes, three recurring design variables explain where results transfer: the validity of the state under a changed input, the cost of deciding whether to reuse it, and the backend's ability to realize the saved work. A feature path tuned for one image checkpoint may need a new refresh schedule on another. Video contexts need a policy for which historical tokens remain relevant. Cross-request caches need evidence that the retrieved trajectory can satisfy a different condition. These factors connect the mechanism to the actual quality and latency trade-off of the deployment.

\section{Model Compression and Quantization}
\label{sec:model-compression}

A denoiser can be expensive because its weights must be moved repeatedly,
because each evaluation executes a large network, or because that network
operates on a long latent sequence. These costs respond to different
interventions. Lower precision changes the representation of numerical
operands. Pruning, factorization, and distillation can change the network
that executes. A more compact latent representation changes the input to
every evaluation and the signal from which the decoder must reconstruct
the output. At fixed sampler, guidance policy, and evaluation count $K$,
the principal latency target in Equation~\ref{eq:latency} is the time $\tau_j$
of each evaluation. Changes to model residency or decoding can also affect
$T_{\mathrm{exposed}}$ or $T_{\mathrm{decode}}$. Peak device memory depends
on which weights, activations, caches, and workspaces coexist.

\begin{figure}[!htbp]
    \centering
    \includegraphics[width=\linewidth]{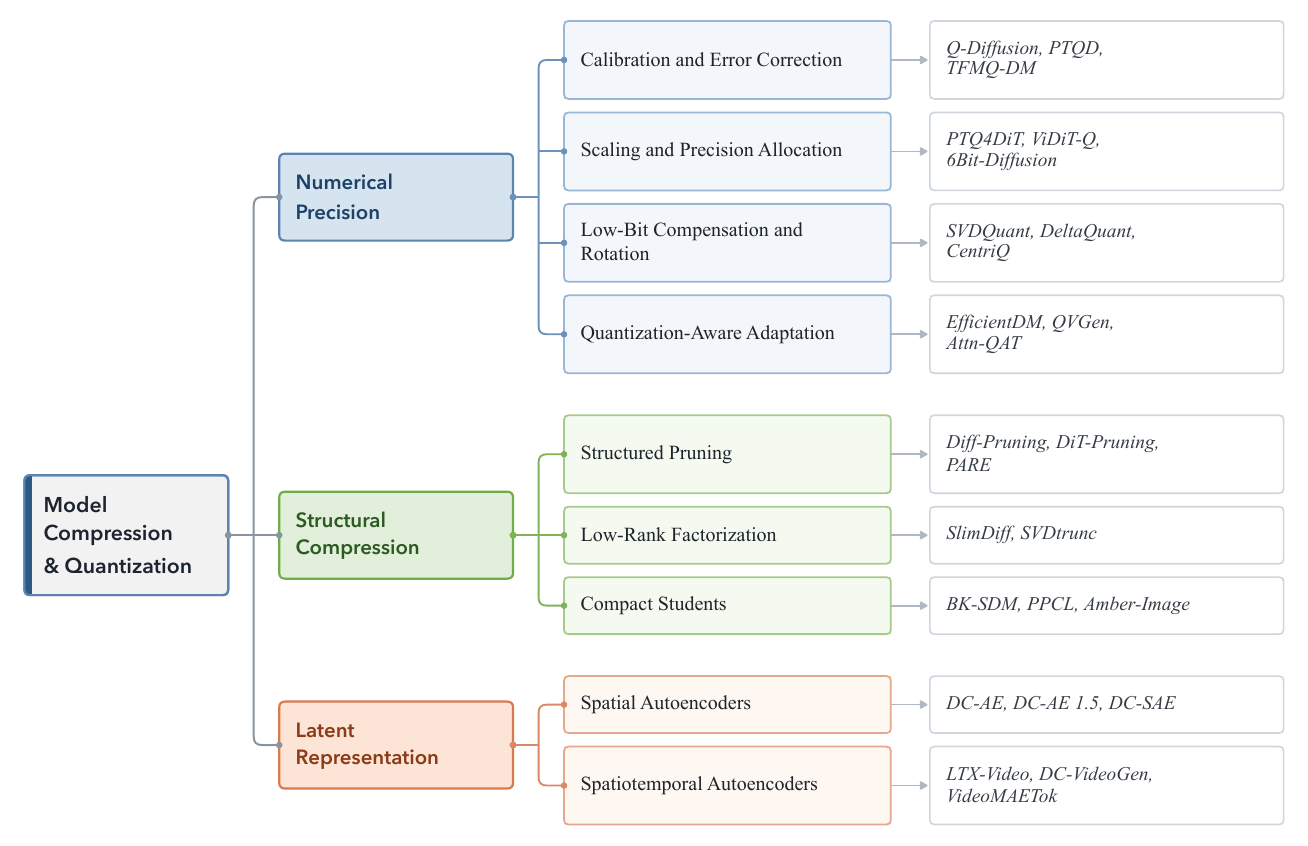}
    \caption{Compression methods grouped by the primary change to
    inference. The nine subtypes distinguish how methods reduce numerical
    precision, change network structure, or shorten the deployed latent
    sequence. The leaves give selected examples\ifsurveyappendices\space from
    Table~\ref{tab:compression-map}\fi.}
    \label{fig:compression-taxonomy}
\end{figure}

\begin{figure}[!htbp]
    \centering
    \includegraphics[width=\linewidth]{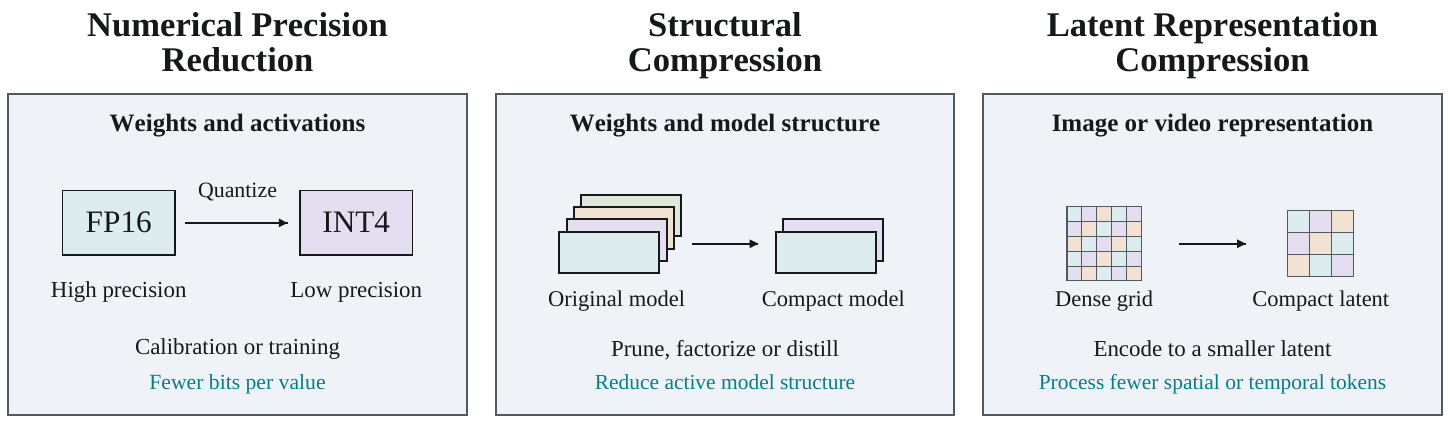}
    \caption{Three objects of compression. Changing the operand format
    affects storage, traffic, and potentially arithmetic. Changing the
    network affects the dimensions, sparsity patterns, or blocks actually
    executed. Changing the latent grid affects every denoiser input and
    the information available to the decoder. FP16 and INT4 are
    illustrative formats.}
    \label{fig:compression-overview}
\end{figure}

Figure~\ref{fig:compression-taxonomy} groups methods by their primary
intervention. Figure~\ref{fig:compression-overview} shows where that
intervention acts in the deployed pipeline.
\ifsurveyappendices
Table~\ref{tab:compression-map} in Appendix~\ref{app:methods:compression}
records the preparation and evaluation setting of each method.
\fi


\subsection{Numerical Precision Reduction}
\label{sec:compression-numerical}
\label{sec:compression-challenges}
\label{sec:compression-ptq}
\label{sec:compression-qat}

For $P$ weights stored uniformly at $b_w$ bits, the packed payload is
approximately $P b_w/8$ bytes before scales, padding, and exceptions.
Weight-only quantization is especially useful when weight reads limit
execution or when packing makes the model resident. A notation such as
W4A4 specifies the widths of weights and activations in the stated
operators. The format, group size, accumulator, and treatment of outliers
determine which kernel can consume them. A compute-limited projection
benefits from native low-bit arithmetic only when its shape and operand
layout suit the kernel and the saved time exceeds conversion and scaling
costs. The projections that produce queries, keys, and values can use
different precision from the attention operation that consumes them.

The same denoiser encounters different activation distributions as noise
is removed. Its output at one evaluation also determines the input at the
next. Guidance adds a second dependence. At a fixed state and noise
level, let $f_{\mathrm u}$ and $f_{\mathrm c}$ denote the reference
unconditional and conditional predictions, and let $e_{\mathrm u}$ and
$e_{\mathrm c}$ denote their signed compression errors. For
classifier-free guidance (CFG) with scale $s$,
\[
  f_s = f_{\mathrm u} + s(f_{\mathrm c} - f_{\mathrm u}),\qquad
  \widehat{f}_s - f_s = e_{\mathrm u} + s(e_{\mathrm c} - e_{\mathrm u}),
\]
which is algebraically equivalent to $(1-s)e_{\mathrm u} + s e_{\mathrm c}$.
An error shared by the branches passes through once. A mismatch between
them is scaled by $s$, even when their separate reconstruction errors
have similar norms. This identity applies to pruning and compact
students as well as quantization whenever both branches are evaluated.
It measures a local error at a common state; after an update, the two
trajectories may visit different states.

\paragraph{Calibration and error correction.}
The choice of calibration target follows from where an error enters the
update. Q-Diffusion samples multiple noise levels and quantizes the two
sides of a U-Net shortcut separately when their activation ranges differ.
TFMQ-DM protects the features that encode the noise level itself. PTQD
instead corrects a component correlated with the denoiser prediction,
adjusts bias, and accounts for residual variance in stochastic
sampling~\cite{li2023qdiffusionquantizingdiffusionmodels,Huang_2024,he2023ptqd}.
Its variance adjustment uses a sampler noise term; a deterministic ODE
update requires a separate treatment. Sampling-Aware Quantization aligns
direction estimates across solver orders and offers both post-training
reconstruction (SA-PTQ) and low-rank adaptation (SA-QLoRA)~\cite{Zeng_2026_CVPR}.
For a few-step solver, the informative comparison fixes the evaluation grid and
NFE and measures the error in the resulting sample.

Calibration across noise levels also applies to audio, but image prompts
and image fidelity do not test the relevant failure modes. PTQ4ADM
selects text conditions and intermediate activations for audio
calibration. On Stable Audio Open, an audio DiT PTQ study compares static
and dynamic scales at W8A8 and W4A8 using objective measures and
listening tests~\cite{vora2024ptq4admposttrainingquantizationefficient,khandelwal2025posttrainingquantizationaudiodiffusion}.
Together these studies distinguish coverage of the model's input states
from the perceptual consequences of an error over the duration of a sound.

\paragraph{Scaling and precision allocation.}
A scale fixed across samples is inexpensive to apply but must cover
channels whose magnitudes can change with the timestep. PTQ4DiT uses
timestep statistics to balance salient channels. Q-DiT searches
quantization granularity and adjusts activation scales by sample, with
evaluations on both image and video DiTs. ViDiT-Q combines channel
balancing with finer scales and allocates precision by
sensitivity~\cite{wu2024ptq4ditposttrainingquantizationdiffusion,Chen_2025,https://doi.org/10.48550/arxiv.2406.02540}.
Online choices incur reductions or dispatch at inference. For example,
6Bit-Diffusion uses a predictor to select NVFP4 or INT8 execution for
video DiT layers (with weight conversion overhead on the fallback path)
and also introduces a temporal delta cache~\cite{su2026sixbitdiffusion}.
Timing the precision policy with the cache disabled isolates a different
saving from timing the full system. Simulated quantization can establish
fidelity, while timing requires the corresponding executable formats,
including fallback and conversion.

\paragraph{Low-bit compensation and rotation.}
Several methods retain a low-bit main multiplication and protect a signal
that would otherwise set its range. SVDQuant carries outliers in a
higher-precision low-rank branch and fuses work across the two paths.
DeltaQuant exploits local correlation in video: it retains an FP8
representative for a space--time neighborhood, codes the smaller token
differences at four bits, and also applies low-rank compensation to
weights~\cite{li2025svdquantabsorbingoutlierslowrank,Li_2026_DeltaQuant}.
The first approach depends on a useful low-rank component. The second
additionally depends on nearby tokens remaining similar. Both leave work
outside the low-bit multiplication.

OrbitQuant and CentriQ reduce dependence on a representative calibration
set by rotating the operands. OrbitQuant uses normalized rotated
coordinates and fixed codebooks. CentriQ first removes the token mean and
restores its contribution through a higher-precision rank-one
path~\cite{lee2026orbitquant,jovanovic2026centriq}. Their deployment cost
therefore includes different transformations and corrections. PulseQuant
addresses a separate limitation of local calibration. It tests how errors
introduced at particular blocks and steps propagate, then adjusts weight
codes along influential response directions while retaining four-bit
weight storage~\cite{wang2026pulsequant}. The stored width identifies the
weight representation; it leaves activation and attention precision to
be specified separately.

\paragraph{Quantization-aware adaptation.}
Recovery methods differ in what can change and what supervision is
available. EfficientDM trains quantization-aware low-rank adapters,
whereas QuEST updates selected sensitive
parameters~\cite{he2024efficientdmefficientquantizationawarefinetuning,wang2025questlowbitdiffusionmodel}.
In video, matching a frame independently can miss an error in its
relation to neighboring frames. Q-VDiT combines token-aware error
estimation with supervision on frame relations. QVGen uses an auxiliary
low-rank module during training and progressively removes it through rank
decay to avoid inference overhead~\cite{https://doi.org/10.48550/arxiv.2505.22167,huang2026qvgen}.
These designs have different teacher, training-memory, and inference
requirements even at the same deployed bit width.

S$^2$Q-VDiT is described by its authors as PTQ, yet its calibration selects
salient examples and weights teacher supervision by token importance,
placing its recovery mechanism alongside adaptation methods\ifsurveyappendices\space in
Table~\ref{tab:compression-map}\fi. In the authors' CogVideoX-5B experiment on
one A800 at batch size one, the floating-point and W4A6 models occupy
10.375 and 2.633\,GB of model storage, respectively. Reported inference
memory falls from 15.801 to 10.145\,GB and latency from 259.2 to 203.2\,s.
On the same W4A6 setup, PTQ4DiT takes 203.1\,s with the same 2.633\,GB
storage, but its imaging-quality score is 43.54 compared with 60.75 for
S$^2$Q-VDiT and 61.80 for the floating-point
baseline~\cite{feng2025s2qvdit}. The floating-point comparison measures the
benefit of the deployed low-bit configuration, whereas the comparison
between the two quantized models isolates the quality recovered at
essentially the same measured latency.

Two further cases identify different limits of adaptation. DSAQuant
changes supervision across denoising stages and disables explicit CFG
in the final steps; its reported gain therefore reflects both the recovered
weights and the guidance policy~\cite{li2026dsaquant}. Attn-QAT targets
FP4 attention itself. It aligns score recomputation in the backward pass
with the low-precision forward pass and treats sensitive gradient
operations separately~\cite{pmlr-v306-zhang26eq}. Its attention results
apply to a different operator from the linear projections in most W4A4
denoisers.

\subsection{Structural Compression}
\label{sec:compression-structure}

For a fixed sampling trajectory, structural compression changes the
capacity or work of an individual denoiser. Three models with the same
reported fraction of retained weights can execute quite differently:
one can use narrower dense operators, another can retain the original
shapes and apply a sparse kernel, and a third can select a different
set of blocks on each input.

\paragraph{Structured pruning and executable sparsity.}
Diff-Pruning estimates importance at informative timesteps and removes
coupled parameter groups so that adjacent layers remain compatible.
DiT-Pruning balances weight and activation information when selecting
weights, including experiments with unstructured and $N{:}M$
masks~\cite{fang2023spd,hu2026ditpruning}. Dense dimensions characterize
the former saving. For a mask, the relevant quantities are the stored
pattern and the work the selected kernel can skip. PARE combines narrower
attention and feed-forward layers with a router that selects blocks
according to content and noise level~\cite{wang2026pare}. Its width can
be recorded once, while executed depth has a distribution over samples
and steps.

The importance criterion also depends on the output. Compact 3DiT
assesses the effect of removing a layer on generated geometry, uses that
information for pruning and precision allocation, and matches both
conditional and unconditional teacher predictions during
recovery~\cite{lee2026compact3dit}. A layer whose removal changes a thin
structure can be important even if a rendered view still looks
plausible, providing direct evidence for geometry-aware recovery in
image-conditioned shape generation.

\paragraph{Low-rank factorization.}
Replacing an $m\times n$ projection by two rank-$r$ factors reduces its
weight count from $mn$ to $r(m+n)$ when $r<mn/(m+n)$, ignoring biases.
An explicit two-kernel implementation writes and reads an intermediate of
width $r$. Rank alignment, sequence length, and fusion determine whether
the smaller arithmetic count yields a shorter forward pass. SlimDiff
uses activation-informed subspaces to reduce coupled attention and
feed-forward dimensions without gradient-based recovery. SVDtrunc
allocates ranks under a global parameter budget and recovers the
factorized student with modular distillation and a rectified-flow
objective~\cite{roy2025slimdiff,zavadski2026svdtrunc}. At 68\% of the
original FLUX.dev parameter count, SVDtrunc-m reports GenEval 0.645
versus 0.647 for the original model, providing a quality comparison at a
specified model size rather than a direct latency claim.

The location of a subspace determines its classification. SVDtrunc
replaces deployed weights. SVDQuant retains a corrective branch beside
a low-bit projection. Video-SVD instead learns bases for approximating
video attention at inference without modifying model weights~\cite{pmlr-v306-wan26g}.
Only the first of these three uses the factors as the structural
replacement for the original projections.

\paragraph{Compact students.}
A smaller student can learn the teacher's prediction at the same noise
level. BK-SDM removes U-Net blocks and distills the remaining network.
PPCL trains depth and width alternatives by distilling contiguous
intervals of a transformer~\cite{kim2024bksdmlightweightfastcheap,Ma_2026_CVPR}.
Amber-Image additionally changes how conditioning reaches the image
stream: its 10B variant reduces depth, while its 6B variant also merges
the later image and text streams~\cite{yang2026amberimage}. A prompt
alignment score can remain high for an image containing a sign with the
wrong lettering; text rendering and binding therefore deserve separate
tests when compressing a text-to-image model's streams.

Step distillation in Section~\ref{sec:sampling-fewstep} changes the
transition that the student must learn. Reducing capacity and asking the
student to cover a longer interval can each affect attainable quality.
Comparing teacher and compact student at matched NFE isolates the change
in capacity. Sweeping NFE for each capacity then estimates the smallest
tested budget that meets a stated quality tolerance. That operating point
depends on the student, supervision, sampler, and task.

\subsection{Latent Representation Compression}
\label{sec:compression-representation}

A tokenizer determines how many visual tokens enter each evaluation and
what the decoder can recover from them. For $T$ frames of size $H\times W$,
let $(s_t,s_h,s_w)$ be the encoder strides and $(p_t,p_h,p_w)$ any
additional denoiser patch strides in latent coordinates. Away from
padding and causal boundary effects, the visual sequence length is
\[
  N_{\mathrm{vis}}\approx
  \frac{T H W}{s_t s_h s_w\,p_t p_h p_w}.
\]
At fixed patching, doubling both spatial strides quarters the visual
token count. The dense visual self-attention interaction term is
quadratic in this count at fixed head width, whereas projection cost,
latent channel width, conditioning tokens, and decoding follow different
scaling. The layout change persists throughout the trajectory; runtime
token selection in Section~\ref{sec:attention-adaptive} instead changes
which tokens participate in selected network operations.

\paragraph{Spatial autoencoders.}
A compact grid must preserve image detail and also provide a distribution
that the denoiser can learn. DC-AE uses residual autoencoding and staged
training to support large spatial strides. DC-AE~1.5 organizes channels
so that structural content and fine detail can receive different
supervision~\cite{chen2025deepcompressionae,Chen_2025_ICCV_DCAE}.
DC-SAE combines a semantic encoder with a pixel-detail
path~\cite{huang2026dcsae}. Its reconstruction PSNR and generative FID
test different parts of the pipeline: the former starts with an observed
image, whereas the latter includes the ability to generate suitable latents
from noise. Two tokenizers at equal stride can consequently exhibit
different generation quality or require different denoiser capacity.

\paragraph{Spatiotemporal autoencoders.}
Temporal stride adds a dependence on motion. LTX-Video designs the
tokenizer and denoiser together, incorporates patchification in the
autoencoder, and assigns a final denoising operation to the decoder.
Its stated pixel-to-token strides are $(H,W,T)=(32,32,8)$. DC-VideoGen
adapts a pretrained video denoiser to chunk-causal latents with spatial
stride 32 or 64 on each axis and temporal stride
4~\cite{hacohen2025ltxvideo,chen2025dc}. The alternatives differ both in
how much motion a token must encode and in the work required to reuse a
checkpoint. VideoMAETok addresses the compatibility of tokenizer and
denoiser through masking and latent corruption during tokenizer
training~\cite{pmlr-v306-tong26i}; the training mask fraction and the
deployed sequence length are separate quantities. At a matched output
frame rate and duration, rapid motion, fine moving textures, and
long-clip drift test information that a frame-averaged reconstruction score
can miss. Timing the decoder is particularly consequential when it also
performs part of the refinement.

\subsection{Transfer and Composition}
\label{sec:compression-discussion}

\paragraph{What transfers across tasks.}
Coverage of the states visited at inference is a transferable calibration
principle, yet both the relevant states and the consequences of a miss
depend on the task. An image model can lose small lettering while
retaining global layout. A video model can preserve individual frames
while distorting their motion. An audio denoiser must preserve
continuity and perceptual detail over the clip, and an image-conditioned
3D model must preserve geometry across views. These observations motivate
task-specific calibration examples and recovery targets.
\ifsurveyappendices Table~\ref{tab:compression-map} contains\else The studies reviewed here include\fi{} direct evaluations in these four
settings, while offering no direct evaluation of closed-loop world
modeling or synchronized joint multimodal generation. In the former, a
compression error can alter the state on which the next action is
conditioned; in the latter, each output stream can remain plausible
while their relative timing drifts. Matched action rollouts and
cross-stream synchronization tests, respectively, are therefore required
to establish whether an image or video compression policy transfers to
those settings.

\paragraph{Guidance and adaptive computation.}
Explicit CFG, distilled guidance, and branch reuse expose different
compression errors. Guided Distillation first learns a single prediction
for the guided teacher and then distills steps, and guidance-free
students such as the SD3-Turbo model in Section~\ref{sec:sampling-fewstep}
also execute without the original pair of
branches~\cite{meng2023distillation,sauer2024fast}.
For these students the reference is the guided target they were trained to
approximate, so further compression perturbs that single prediction. By
contrast, FasterCache estimates a guidance contribution from reused
information, and DiTFastAttn shares selected attention computation across
CFG branches~\cite{lv2024fastercache,yuan2024ditfastattn}. Branch
differences measured on the original model can change after pruning or
quantization. For these policies, compare fresh and reused branches on
the same compressed checkpoint. For DSAQuant, compare the original and
gated guidance schedules on both the reference and quantized models:
changing $s$ to $\hat{s}$ alone shifts the full-precision target by
$(\hat{s}-s)(f_{\mathrm c}-f_{\mathrm u})$ at a fixed state. Evaluations
should record branch evaluations as well as sampler steps, including
whether the two branches share a batched invocation.

Precision can also change a discrete decision downstream. A top-$p$
block selector can cross its cumulative-mass cutoff after a small change
to attention scores, a token reducer can reverse two nearly tied saliency
ranks, and a cache predictor can cross a refresh threshold even when the
mean feature error is small. Co-designed methods such as PAROAttention
and SLA2 therefore align pattern reordering or sparse--linear routing
with low-bit execution and quantization-aware fine-tuning, rather than
stacking independent approximations~\cite{zhao2025paroattention,zhang2026sla2}
(Sections~\ref{sec:attention-adaptive} and~\ref{sec:attention-linear}).
On identical states, mask overlap at matched density, changes to retained
tokens, and refresh-decision disagreement locate these effects; running
the resulting policies along their own trajectories then measures the
quality and latency consequences. Historical KV quantization, as in Quant
VideoGen, changes the stored history read by later
queries~\cite{xi2026quantvideogenautoregressivelong}, whose capacity and
read traffic are treated in Section~\ref{sec:caching:attention}. In
masked language diffusion, the analogous sensitive decision is which
token to commit or when to stop, so Section~\ref{sec:dllm:adaptation}
evaluates low-bit execution on evolving mask states and token margins.

\paragraph{Which savings reach the user.}
Once linear projections become faster, attention, normalization,
rotations, or format conversion can occupy a larger fraction of a
denoiser evaluation. Text encoding ($T_{\mathrm{condition}}$) or VAE
decoding ($T_{\mathrm{decode}}$) can then limit end-to-end gains even
when every denoising evaluation speeds up. Conversely, a smaller model
can eliminate host--device transfers that were previously on the critical
path ($T_{\mathrm{exposed}}$). Sol-Engine and TurboDiffusion combine
numerical changes with other mechanisms, while Diffusers group offload
and ChunkFlow change residency and exposed transfer
time~\cite{solengine2026,turbodiffusion2025,diffusersmemorydocs,chunkflow2026}
(Section~\ref{sec:systems}). A resident-to-resident comparison isolates the
arithmetic and traffic savings where both configurations fit, whereas a
fixed-device-budget comparison captures the additional benefit when only
the compressed model fits.

For a combined method, compare the reference, each intervention alone,
and the combination at the same workload and task-quality tolerance.
Record the checkpoint and preparation cost, sampler and branch-counting
convention, output shape, batch size, device, kernel, and offload policy.
Model storage and peak device memory answer different capacity questions,
and operator time, summed denoiser time, and request latency show where a
saving enters Equation~\ref{eq:latency} and how much of it remains on the
critical path.

\section{System Optimization and Parallel Execution}
\label{sec:systems}

\begin{figure}[!htbp]
  \centering
  \includegraphics[width=\linewidth]{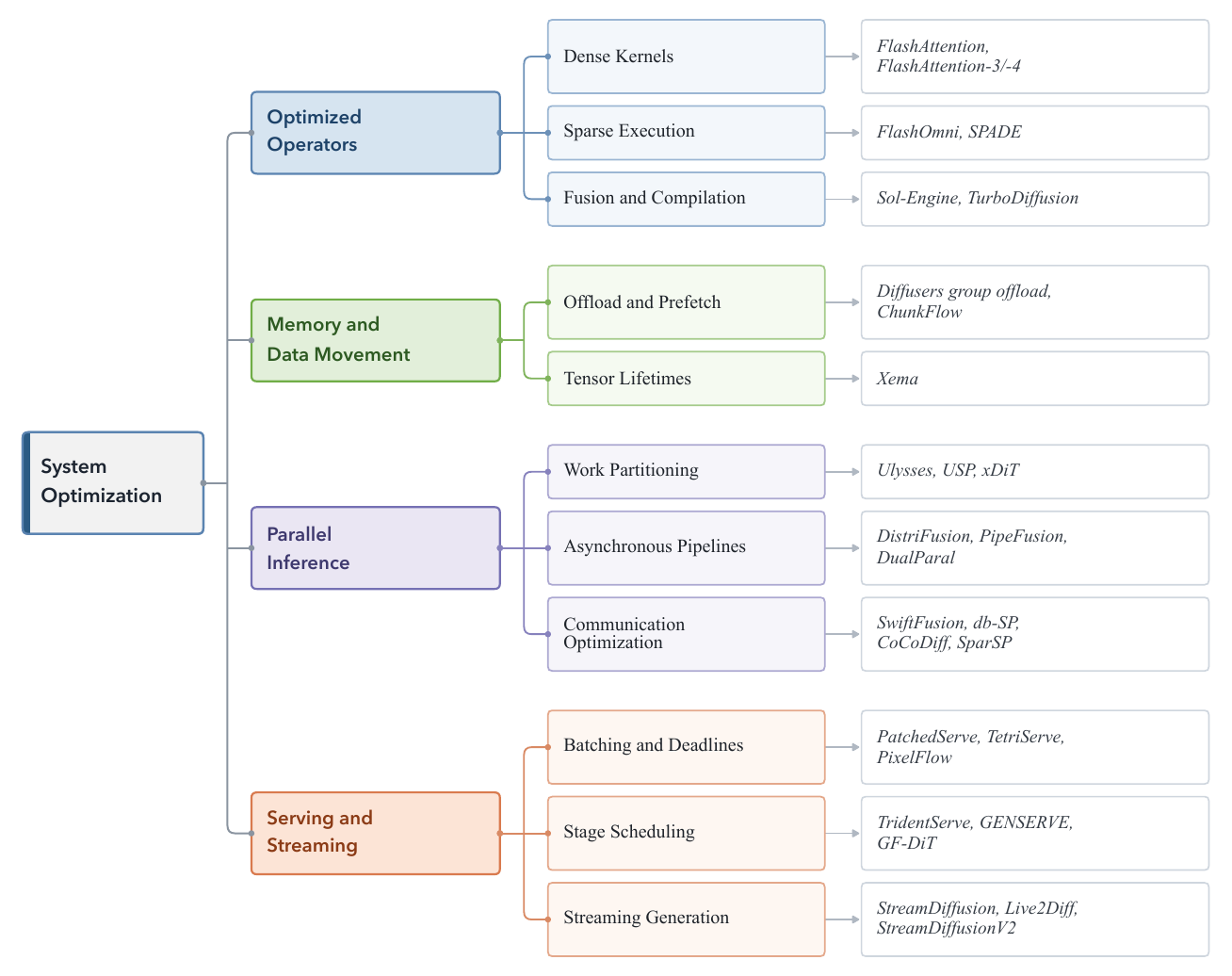}
  \caption{Taxonomy of system optimization for diffusion inference. Leaves show examples by primary mechanism.\ifsurveyappendices\space Table~\ref{tab:systems-methods} provides the complete inventory.\fi}
  \label{fig:systems-taxonomy}
\end{figure}

\begin{figure}[!htbp]
    \centering
    \includegraphics[width=\linewidth]{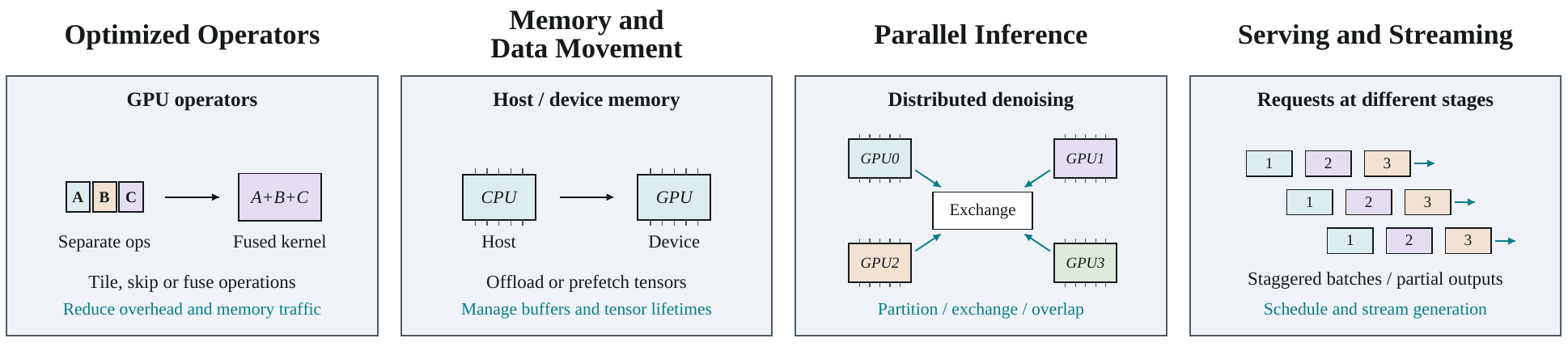}
    \caption{System optimization for diffusion inference. Optimized operators tile, skip, or fuse computation. Memory management coordinates tensor lifetimes and transfers. Parallel inference partitions work across devices and exchanges state. Serving and streaming schedule requests and release outputs incrementally. The exchange block denotes logical communication among GPUs.}
    \label{fig:systems-overview}
\end{figure}

Diffusion inference repeatedly moves weights, activations, and conditioning information through a sequence of denoising evaluations. A systems optimization changes how these computations are executed or scheduled. Its impact depends on which resource limits the workload: GPU arithmetic, memory capacity or bandwidth, interconnect traffic, or the time requests spend waiting to run. Diffusion serving adds variation in image size, video duration, denoising budget, and deadlines. \ifsurveyappendices Table~\ref{tab:systems-methods} in Appendix~\ref{app:methods:systems} lists representative methods and their evaluation scope, and \fi{}Figure~\ref{fig:systems-taxonomy} organizes them by their primary intervention. The discussion centers on visual diffusion pipelines, with diffusion language model systems covered in Section~\ref{sec:dllm}. Figure~\ref{fig:systems-overview} follows the execution path from kernel and memory optimization to device-level parallelism and request scheduling.

\subsection{Optimized Operators}

\subsubsection{Dense Kernels}
FlashAttention tiles the dense attention calculation and applies online softmax to reduce transfers between on-chip storage and high-bandwidth memory, retaining the full query--key interaction pattern~\citep{dao2022flashattention}. FlashAttention-2 improves utilization by repartitioning work across thread blocks and warps~\citep{flashattention22023}. FlashAttention-3 targets Hopper with asynchronous pipelining and a separate low-precision path~\citep{flashattention32024}. FlashAttention-4 adapts the kernel to Blackwell, where matrix throughput and the throughput of operations such as softmax scale differently~\citep{flashattention42026}. These designs illustrate how bottlenecks shift across GPU generations. Kernel comparisons should identify the backend, mask, input shape, and precision, since the best strategy for an individual attention operation can change with each of these settings.

\subsubsection{Sparse Execution}
Reducing the interaction count yields a speedup when the remaining work maps to sufficiently large, balanced GPU blocks. FlashOmni provides a unified execution path for multiple sparse layouts and also targets projection operations~\citep{qiao2025flashomniunifiedsparseattention}. SPADE generates head-specific execution schemes and groups block-sparse kernels, accounting for indexing and layout overhead~\citep{spade2026}. Its reported attention speedup exceeds its end-to-end speedup, reflecting the time still spent outside the optimized operator. Section~\ref{sec:attention} discusses how masks and sparse patterns are chosen. Here the comparison centers on turning a given pattern into a runnable workload.

\subsubsection{Fusion and Compilation}
Fusing adjacent operations can avoid intermediate memory writes. Graph capture can amortize host launch overhead when shapes and execution paths are stable, while dynamic resolution, step counts, or sparse layouts may require multiple compiled paths. Full-stack engines put these mechanisms alongside algorithmic changes. Sol-Engine selects combinations of reuse, sparsity, token pruning, quantization, and fusion for a deployment configuration~\citep{solengine2026}. TurboDiffusion combines low-bit and sparse attention, W8A8 quantization, and step distillation~\citep{turbodiffusion2025}. These combinations call for separate reporting of the full recipe and component ablations. Sections~\ref{sec:sampling} and~\ref{sec:model-compression} cover the step-count and precision changes.

\subsection{Memory and Data Movement}

\subsubsection{Offload and Prefetch}
Offloading enables a larger model or input shape to fit a fixed GPU budget by moving inactive weights or components to host memory. Group offload in Diffusers exposes the granularity and, for supported configurations, allows asynchronous transfer on a separate stream~\citep{diffusersmemorydocs}. Smaller groups usually reduce residency while increasing the frequency of transfers. The relevant latency is the part of those transfers exposed on the critical path. ChunkFlow co-schedules weight prefetch and cross-GPU collectives: on a PCIe node the two operations may contend for the same bandwidth, so chunked transfers can yield to communication~\citep{chunkflow2026}. An evaluation should compare peak device and host memory, exposed transfer time, and the largest feasible shape against both no offload and a conventional offload baseline.

\subsubsection{Tensor Lifetimes}
Peak memory depends on when allocations overlap, as well as on the size of each allocation. Xema profiles tensor lifetimes, constructs a layout for predictable tensors, and applies memory mitigations only during high-pressure intervals~\citep{xema2026}. This is especially relevant for pipelines in which the denoiser and VAE have different temporary footprints. Peak allocated and reserved memory capture different effects of lifetime and fragmentation. Buffer placement and reuse manage storage, whereas the semantic validity of reusing a computed state is discussed in Section~\ref{sec:caching}.

\subsection{Parallel Inference}

\subsubsection{Work Partitioning}
Replicating a model across devices increases the number of independent requests it can process. Reducing the latency or per-device memory of one evaluation requires partitioning its work, for example across weights, tokens, patches, layers, or classifier-free guidance (CFG) branches. Ulysses exchanges sequence and attention-head partitions using all-to-all communication~\citep{deepspeedulysses2023}. Ring Attention circulates key and value blocks so each device can incrementally evaluate attention for its local queries~\citep{ringattention2023}. USP combines the two along separate parallel dimensions~\citep{usp2024}, and xDiT provides several partitioning mechanisms for diffusion transformers, including patch pipelines and CFG parallelism~\citep{xdit2024}. A useful comparison identifies which tensors move, how much work each device retains, and how the mapping relates to interconnect bandwidth and head count.

\subsubsection{Asynchronous Pipelines}
DistriFusion assigns spatial patches to devices and uses previous-step context while communication supplies updated activations~\citep{li2024distrifusion}. PipeFusion pipelines patches across groups of model layers, also exploiting displacement across denoising steps~\citep{pipefusion2024}. DualParal pipelines frame blocks across model layers with a block-wise denoising schedule for long sequences~\citep{dualparal2025}. In such pipelines, asynchronous execution can overlap communication, but uneven stage times or synchronization boundaries limit the overlap. When a method uses stale context or modifies the trajectory, quality should be measured at the targeted resolution and duration against the synchronous or unmodified baseline.

\subsubsection{Communication Optimization}
Topology matters because intra-node and inter-node links can offer different bandwidth and synchronization costs. SwiftFusion combines topology-aware sequence parallelism, Torus Attention, and one-sided communication~\citep{swiftfusion2026}. AoiZora selects an inference configuration using a topology-aware performance model~\citep{aoizora2026}. CoCoDiff overlaps value-tensor exchanges with other projections and also studies selective communication; these are separate sources of gain~\citep{cocodiff2026}. Sparse attention can further change which transfers are necessary and where load accumulates. DSA co-designs sparse attention and distributed execution, db-SP addresses imbalance across partitions, and SparSP uses sparsity to avoid unnecessary exchanges~\citep{li2026dsa,dbsp2026,sparsp2026}. Report communication volume, exposed communication time, and the slowest-device load alongside latency. Single-request acceleration obtained by adding devices should also be evaluated in GPU-seconds.

\subsection{Serving and Streaming}

\subsubsection{Batching and Deadlines}
Requests arriving at different times can share a batch if the scheduler can manage differences in resolution and remaining work. DiffServe routes text-to-image requests through model cascades and adapts resource allocation to demand~\citep{diffserve2025}. PatchedServe batches patches across resolutions, while PixelFlow uses tokens as the unit of distributed workload packing~\citep{patchedserve2025,pixelflow2026}. TetriServe adapts sequence-parallel degree at denoising-step boundaries to meet deadlines~\citep{tetriserve2025}. FlashDiff bypasses low-impact regional updates and schedules remaining regions by affinity~\citep{flashdiff2026}. The latter changes the computation as well as its schedule, so latency and quality need to be compared together. For all systems, report total completed requests per unit time and deadline goodput under a specified arrival process and workload mix.

\subsubsection{Stage Scheduling}
Text encoding, denoising, and decoding have different compute and memory profiles. DDiT independently allocates resources for the DiT and VAE phases and can change the allocation during denoising~\citep{ddit2025}. TridentServe jointly considers stage placement and request dispatch~\citep{tridentserve2025}. GENSERVE jointly serves image and video jobs using video preemption, elastic sequence parallelism, and dynamic batching~\citep{genserve2026}. GF-DiT makes trajectory tasks independently schedulable and forms device groups at runtime~\citep{gfdit2026}. vLLM-Omni provides a runtime that coordinates heterogeneous stages and transfers intermediate payloads~\citep{vllmomni2026}. The benefit of decoupling depends on whether better resource utilization offsets transfer and queueing costs between stages. Benchmarks should include both stage-level utilization and the end-to-end latency distribution.

\subsubsection{Streaming Generation}
Streaming adds a requirement on when each intermediate output becomes usable. StreamDiffusion pipelines denoising stages through stream batches for interactive generation~\citep{streamdiffusion2023}. Live2Diff deals with the video-specific dependency by using causal temporal attention~\citep{xing2024live2difflivestreamtranslation}. StreamDiffusionV2 combines deadline-aware batching and pipelined execution with a rolling context for live video diffusion~\citep{streamdiffusionv22026}. First-frame latency, steady-state FPS, and jitter measure different aspects of responsiveness. For interactive video, deadline misses and temporal consistency should accompany throughput, particularly when comparing systems that expose partial outputs at different granularities.

\subsection{Summary}

An evaluation should separate time spent waiting, encoding inputs, executing denoising steps, and decoding outputs. Attribute communication and data transfers to their exposed time along this critical path. Report output shape, denoising evaluation count, precision, GPU type and count, interconnect, and workload conditions. Evaluate approximate schedules together with their quality effects. For online workloads, distinguish median and tail latency, total throughput, and the proportion or rate of requests that satisfy their deadlines. This accounting makes it possible to compare operator gains, cross-device speedups, and serving improvements without attributing them to the same mechanism.

\section{Efficient Inference for Diffusion Language Models}
\label{sec:dllm}

\begin{figure}[!htbp]
    \centering
    \includegraphics[width=\linewidth]{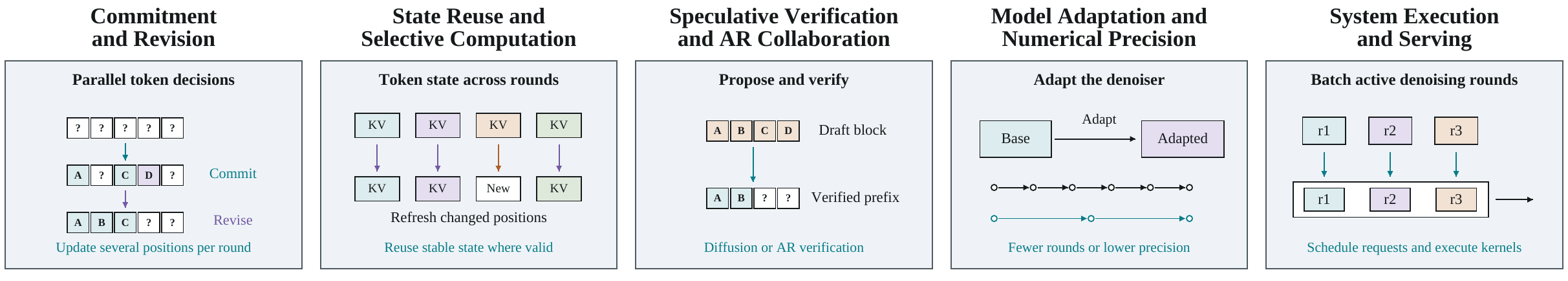}
    \caption{Mechanisms for efficient diffusion language model inference. From left to right, the panels show parallel commitment and revision, selective refresh of token state, speculative verification with diffusion or autoregressive models, model adaptation for fewer rounds or lower precision, and scheduling and kernel execution. A question mark denotes an unresolved token.}
    \label{fig:dllm-overview}
\end{figure}

Masked diffusion language models (dLLMs) repeatedly predict unresolved token positions. Parallel predictions save time when the decoder can commit multiple tokens reliably and reuse computation as the sequence changes. \ifsurveyappendices Table~\ref{tab:dllm:methods} in Appendix~\ref{app:methods:dllm} and \fi{}Figure~\ref{fig:dllm-taxonomy} \ifsurveyappendices organize\else organizes\fi{} acceleration methods for discrete masked and block-diffusion models. Related surveys cover architectures, training, and inference-time scaling in greater breadth~\citep{lin2026efficientdllm,gwak2026maskeddiffusionsurvey}. Figure~\ref{fig:dllm-overview} links the decoding decisions made in each denoising round to model adaptation and serving optimizations.

\begin{figure}[!htbp]
  \centering
  \includegraphics[width=.89\linewidth]{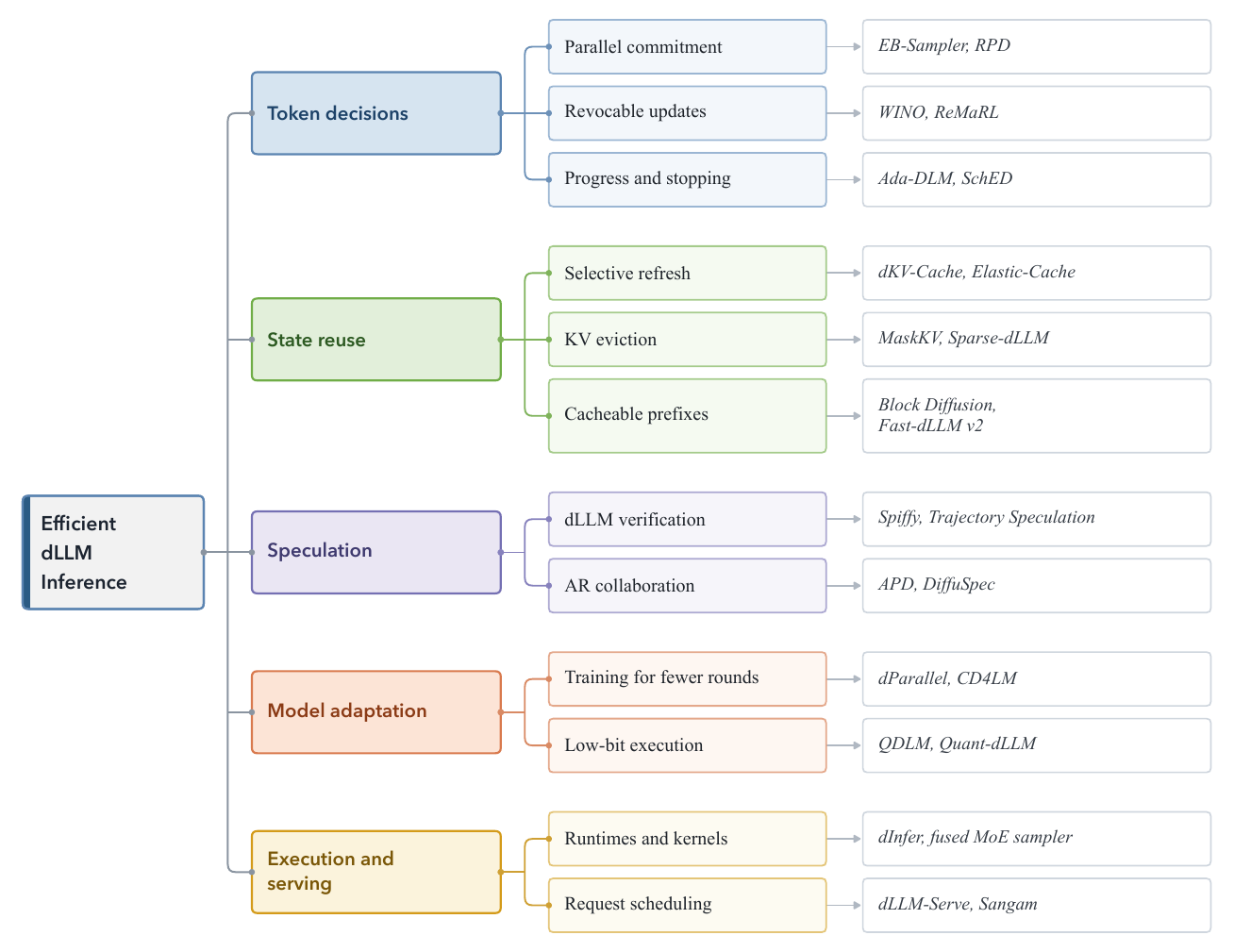}
  \caption{Taxonomy of efficient dLLM inference. The leaves show representative examples.\ifsurveyappendices\space Table~\ref{tab:dllm:methods} gives the method inventory.\fi}
  \label{fig:dllm-taxonomy}
\end{figure}

\subsection{Generation Characteristics and Efficiency Bottlenecks}
\label{sec:dllm:bottlenecks}

\subsubsection{Denoising rounds and revision}
\label{sec:dllm:iterations}

Let $x^{(t)}$ be a partially visible sequence and $\mathcal{M}_t$ its set of unresolved positions. A denoiser evaluates predictions at positions in $\mathcal{M}_t$, and a policy chooses which tokens to commit. Irreversible policies shrink the mask set monotonically. Revocable policies can reopen visible positions when later context changes their plausibility. Forward-pass count measures how often the denoiser is invoked. Work per pass depends on active positions, retained context, and the implementation of attention and sampling.

\subsubsection{Parallel prediction and commitment}
\label{sec:dllm:commitment}

Let $c$ denote currently visible context. The joint conditional distribution of unresolved tokens generally differs from the product of their marginals,
\begin{equation}
  p(x_i,x_j\mid c) \not\equiv p(x_i\mid c)\,p(x_j\mid c).
  \label{eq:dllm:independence}
\end{equation}
For example, a confident final answer may be committed before the intermediate calculations supporting it are settled. Uncertainty-based selection, dependency-aware ordering, and verification make different use of the same parallel predictions. Their output quality can be judged by task scores, agreement with a specified greedy decoder, or preservation of a specified stochastic decoder's output distribution. These criteria answer different questions.

\subsubsection{Bidirectional states and cache validity}
\label{sec:dllm:states}

In causal autoregressive (AR) decoding, appending tokens leaves previously computed KV states valid. Under bidirectional attention, revealing a masked token can change representations at other positions, including positions with fixed token identities~\citep{ma2025dkvcachecachediffusionlanguage,liu2025dllm}. A cache can reuse states approximately by tolerating drift, or it can exploit a dependency structure that prevents such changes. Block diffusion permits exact prefix reuse under its inter-block attention pattern, while states inside the active block can still evolve~\citep{arriola2025blockdiffusioninterpolatingautoregressive}.

\subsection{Accelerating the Denoising Process}
\label{sec:dllm:acceleration}

\subsubsection{Commitment and Revision}
\label{sec:dllm:decisions}

\paragraph{Parallel commitment.}
Fast-dLLM commits multiple confident tokens and combines this policy with approximate KV reuse~\citep{wu2025fastdllmtrainingfreeaccelerationdiffusion}. EB-Sampler selects a set of tokens subject to a joint budget over their marginal entropies, linking the width of a commitment to an approximation bound~\citep{benhamu2025acceleratedsamplingmaskeddiffusion}. DOS uses attention to guide the update order, and RWS reweights token predictions using rewards~\citep{zhou2026dosdependencyorientedsamplermasked,gwak2025rewardweightedsamplingenhancingnonautoregressive}. Reliable Parallel Decoding (RPD) uses stability across late layers and final-layer confidence to screen predictions. Its entropy budget is applied to preceding unresolved positions to defer predictions whose upstream context is uncertain~\citep{he2026reliableparallel}. Intermediate-logit extraction, reward scoring, and dependency selection are part of the inference budget.

\paragraph{Revocable updates.}
WINO lets a decoder commit a wide set and subsequently remask tokens when their support weakens~\citep{hong2025wideinnarrowoutrevokabledecoding}. A reassessment found that confidence-based remasking brought little additional benefit at shorter block lengths under greedy decoding, with gains appearing at longer block lengths in the tested settings~\citep{frkovic2026confidence}. ReMaRL trains a small controller with final-answer rewards to unmask, retain, or remask positions while freezing the denoiser~\citep{shibaev2026remarl}. Its reported improvements compare task quality at matched model-call budgets. Runtime claims additionally require measuring controller overhead and actual revision work.

\paragraph{Progress and stopping.}
Ada-DLM tracks confidence histories and adaptively groups positions to identify states that need further updates. Its intervention is position-specific~\citep{li-etal-2026-towards-efficient}. SchED instead uses a progress-dependent logit-margin rule to stop a sequence early~\citep{mohamed2025fastdecodingdiffusionlanguagemodels}. Both reduce repeated work, with different consequences for the remaining masked positions and for output length.

\subsubsection{State Reuse and Selective Computation}
\label{sec:dllm:caching}

\paragraph{Selective refresh.}
dKV-Cache exploits the evolution of token representations to delay KV reuse until states become more stable~\citep{ma2025dkvcachecachediffusionlanguage}. dLLM-Cache uses different refresh policies for prompt and response states, including similarity-based updates for the response~\citep{liu2025dllm}. d$^2$Cache combines certainty and attention to choose refresh targets~\citep{jiang2026d2cacheacceleratingdiffusionbasedllms}. Elastic-Cache checks drift at a highly attended token and selectively recomputes deeper layers while reusing shallow-layer and off-window states~\citep{nguyentri2026elasticcache}. These policies trade recomputation for drift, which can shift the logits used by subsequent commitment rules.

\paragraph{KV eviction.}
MaskKV uses attention from unresolved positions to estimate which prompt states will be useful, then allocates a KV budget across layers and heads~\citep{huang2025masktokensprophetfinegrained}. Sparse-dLLM exploits the persistence of context-token salience across rounds to evict less useful states and execute sparse attention~\citep{song2025sparsedllmacceleratingdiffusionllms}. Refresh asks whether a stored representation is current enough to reuse. Eviction asks whether the representation should remain available at all. A smaller cache reduces end-to-end latency only to the extent that sparse execution and lower memory traffic offset selection and indexing overhead.

\paragraph{Cacheable prefixes.}
Block Diffusion imposes an ordered dependency across blocks and retains bidirectional context within a block~\citep{arriola2025blockdiffusioninterpolatingautoregressive}. Fast-dLLM v2 uses fine-tuning to adapt an AR checkpoint to block diffusion, then combines a completed-prefix cache with sub-block caching during partial decoding~\citep{wu2025fastdllmvtwo}. Prefix cache validity follows from the inter-block dependency structure. Reuse inside the active block requires its own justification. Block size also controls the trade-off between available parallelism and the number of unresolved dependencies.

\subsubsection{Speculative Verification and AR Collaboration}
\label{sec:dllm:verification}

\paragraph{Verification with a diffusion target.}
Spiffy builds calibrated draft graphs, and SSD organizes candidate updates for hierarchical self-verification~\citep{agrawal2026structuringfuturediffusionllm,gao2025selfspeculativedecodingdiffusion}. Trajectory-level speculation drafts sequences of update sets and verifies them in parallel with bidirectional attention masks. Its exactness result is conditional on the trajectory construction and verification assumptions. In the authors' experiments, the method achieved a $1.3\times$ speedup over Fast-dLLM, the accelerated baseline built on the same dual-cache infrastructure~\citep{pan2026trajectory}. Flash-dLLM joins a fused KV path with cache-assisted self-drafting and verification, illustrating why verification cost includes memory movement as well as denoiser calls~\citep{nguyentri2026flashdllm}. For all four approaches, evaluation should name the reference decoder and include the cost of rejected candidates, masks, and cache maintenance.

\paragraph{AR collaboration in two directions.}
APD uses a small AR model to add dependency information to diffusion marginals. The decoder samples from a multiplicative mixture and adapts parallel progress through coupled sampling~\citep{israel2025apd}. DiffuSpec uses a dLLM to propose tokens for an AR target and searches for causally consistent candidate paths~\citep{li2025diffuspecunlockingdiffusionlanguage}. APD thus changes how diffusion-generated tokens are selected, while DiffuSpec obtains speedups for AR generation under the target's verification procedure. The two configurations require different baselines and fidelity checks.

\subsection{Model Adaptation and Numerical Precision}
\label{sec:dllm:adaptation}

\paragraph{Training for fewer rounds.}
dParallel trains the model to make sufficiently certain predictions at more positions per pass. CD4LM pairs discrete consistency distillation with adaptive decoding to shorten the inference trajectory~\citep{chen2025dparallellearnableparalleldecoding,liang2026cd4lmconsistencydistillationadaptive}. Comparisons should identify the checkpoint and teacher, the training or fine-tuning cost, and the step budget used at evaluation. Checkpoint changes can shift the quality--latency frontier and the operating point of later cache or commitment policies.

\paragraph{Low-bit execution.}
QDLM studies post-training quantization (PTQ) across bitwidths, activation outliers, models, and tasks. Quant-dLLM calibrates with evolving mask states and uses mixed precision for highly compressed weights~\citep{lin2026quantizationmeetsdllmssystematic,zhang2025quantdllmposttrainingextremelowbit}. Quantization can change logit margins enough to reorder commitments or trigger a different exit decision. Report the realized kernel and precision, peak memory, and measured decoding performance together with task quality. Weight compression and end-to-end acceleration depend on the implementation and workload.

\subsection{System Execution and Serving}
\label{sec:dllm:serving}

\paragraph{Runtimes and kernels.}
dInfer decomposes the engine into the model, iteration manager, decoding strategy, and KV manager, supporting compilation, graph capture, and execution optimizations~\citep{ma2025dinferefficientinferenceframework}. The per-round sampler can itself become a bottleneck. On a quantized diffusion MoE running on NVIDIA GB10, \citet{samrat2026hidden} profile logits sampling and stopping as a major share of decode time and fuse that work. Their study reports distinct bit-exact and benchmark-accuracy-parity configurations, and finds a much smaller sampler share on the tested dense model. Kernel savings therefore depend on how much of decode time the sampler occupies after backbone optimization.

\paragraph{Request scheduling.}
dLLM-Serve addresses memory peaks and schedules cache refresh and reused decode work across requests~\citep{fan2026tamingmemoryfootprintcrisis}. Sangam treats repeated full refreshes and cache-assisted passes as recurring prefill and decode work. Its deficit-budget scheduler prioritizes in-flight decodes and admits a full refresh when the accumulated budget allows~\citep{kedia2026sangam}. On the reported LLaDA-8B ShareGPT workload, colocated serving reduced mean latency by 9--20\% relative to hybrid execution. On the Dream-7B arXiv workload, hybrid execution reduced mean latency by 8--20\% relative to colocated serving. A serving comparison should describe both the request mix and the fixed hardware budget, and report tail latency alongside throughput.

\subsection{Summary and Outlook}
\label{sec:dllm:transfer}

Two runs with the same denoising pass count can achieve different user-visible latency. One may repeatedly refresh a long context, while another spends time evaluating candidate branches or sampling from a large logits tensor. Conversely, fewer model calls can come from wider commitment, a shorter output, or a changed stopping rule. Matched prompt and output budgets and separate reporting of model calls, committed tokens per pass, time to first emitted token, end-to-end latency, throughput, and peak memory help distinguish these mechanisms. Streaming evaluations should also state when a position becomes safe to emit, particularly when the decoder can revise previous tokens.

The general principles of Sections~\ref{sec:sampling}--\ref{sec:systems} apply to text, with discrete commitments and mutable bidirectional states shaping where savings are possible. Long-context and multimodal generation add heterogeneous prompt states and change the balance between refresh and decode. The most informative comparisons hold checkpoint, precision, hardware, and generation policy fixed where possible, then measure the joint effect of interventions on speed, quality, and reference-decoder fidelity. Section~\ref{sec:future} discusses the broader problem of evaluating such composite acceleration methods.

\section{Open Problems and Future Directions}
\label{sec:future}

Table~\ref{tab:cross-task-synthesis} answers the three questions raised in Section~\ref{sec:landscape} using cases from the technical chapters. An acceleration mechanism transfers when its assumptions about state and dependencies hold in the new task. Its measured benefit also depends on which part of the pipeline it accelerates. These findings motivate the following open problems.

\begin{table}[!htbp]
\centering
\small
\caption{Cross-task findings from the acceleration mechanisms surveyed in Sections~4--9.}
\label{tab:cross-task-synthesis}
\setlength{\tabcolsep}{5pt}
\renewcommand{\arraystretch}{1.12}
\begin{tabular*}{\linewidth}{@{\extracolsep{\fill}}lll@{}}
\toprule
\textbf{Question} & \textbf{Evidence} & \textbf{Finding} \\
\midrule
What transfers? & Feature reuse and KV reuse & Cached state must stay valid. \\
What must survive? & Few-step images, streaming video & Video must preserve motion. \\
What reaches users? & SLA: $13.7\times$ kernel, $2.2\times$ end to end & Gains depend on the critical path. \\
\bottomrule
\end{tabular*}
\end{table}

The reuse cases are image feature caching, video history KV, and dLLM active-block KV (Sections~\ref{sec:caching} and \ref{sec:dllm}).
For streaming video, measure first-output delay and steady-state frame rate (Sections~\ref{sec:sampling} and \ref{sec:systems}).
The SLA example comes from Wan2.1-1.3B~\citep{zhang2026sla}.

\subsection{Joint Design of Sampling, Reuse, and Precision}

The sampling methods in Section~\ref{sec:sampling} change both the
number and placement of model evaluations. A shorter schedule can
increase the interval between cache refreshes, changing the error of
feature reuse or extrapolation (Section~\ref{sec:caching}).
Timestep-dependent quantization adds another interaction
(Section~\ref{sec:model-compression}): a new schedule changes the
set of states encountered at inference, and quantization can change
the scores that adaptive attention uses to select tokens or blocks.
An open problem is how to allocate computation jointly across
sampling steps, cache refreshes, and numerical precision.
A controlled study could compare each intervention alone, their
independently tuned combination, and a jointly tuned pipeline at
the same task-quality tolerance. Measuring latency and intermediate
error across the trajectory would help locate where the components
interact and whether those interactions shift the best operating
point.

\subsection{Transfer across Tasks and Representations}

The mechanism--task relationships in Section~\ref{sec:landscape}
identify where the reviewed methods have been applied. They provide
starting points for testing transfer across representations.
For example, an image feature cache exploits redundancy across
denoising steps. Its use in video has to account for temporal
dependencies, including motion and the recurrence of previously
observed content. In~3D generation, a staged pipeline such as TRELLIS
allocates denoising work across structure and appearance, so savings
in one stage must be assessed alongside geometry quality and the
remaining reconstruction or decoding cost
(Section~\ref{sec:sampling}). Similarly, the video latent
compression methods in Section~\ref{sec:model-compression} reduce
the number of tokens processed by the denoiser, while their temporal
stride affects the motion detail available to the decoder.
Transfer studies should compare the original policy and its
target-adapted version against a matched baseline on the target
task. Reporting calibration, retraining, and kernel changes would
make it possible to judge how much of the original benefit survives
and what adaptation is required to recover it.

\subsection{Realized Efficiency under Long and Variable Workloads}

The attention and systems chapters show why workload and
implementation matter together. Sparse operators with the same
attention density can incur different routing, memory-access, and
communication costs
(Sections~\ref{sec:attention} and~\ref{sec:systems}).
Cross-request caching introduces a different source of variability:
the overall benefit depends on the hit rate, the cost of probing and retrieving compatible state, and the cost of fallback after a miss
(Section~\ref{sec:caching:content}). These mechanisms motivate
evaluating complete pipelines across workload mixes and arrival
rates. Under a fixed device budget, measurements should track
end-to-end latency, exposed communication time, and goodput,
measured as the rate of requests completed within their deadlines.

Long-running generation adds the question of which states to retain
and when to refresh them. A rolling video context must preserve
information across scene changes and revisits. In world models,
errors can propagate through later interactions, and joint
multimodal generation also requires synchronization across streams.
A useful test would replay sequences with steady intervals, abrupt
changes, and delayed revisits under matched memory budgets.
First-output delay, tail latency, and long-range consistency should
be measured on the same sequences, so that a faster refresh policy
can be evaluated against the dependencies it must preserve.

\subsection{Reliable Parallel Progress in Continuous and Discrete Diffusion}

Both continuous and discrete diffusion can save computation by
concentrating evaluations on states that still require refinement.
Section~\ref{sec:dllm} shows that the unit of useful progress in
masked language generation includes the tokens committed per pass.
Those commitments condition subsequent predictions, and the
representations of already committed tokens can still change under
bidirectional attention. Block diffusion exposes a structural
boundary: completed prefixes can be cached under the inter-block
attention pattern, while reuse within the active block requires its
own validity criterion. Verification introduces a further trade-off
between useful accepted work and the cost of checking candidates.

A concrete question is whether stability signals can predict the
benefit of additional computation across these settings. In
continuous diffusion, this can be tested by measuring how skipped
updates or reused states affect the trajectory and final task
quality. In dLLMs, the corresponding test should include committed
tokens per pass, refresh and verification costs, and the point at
which an output token can be emitted. Task-quality parity, agreement
with a specified deterministic decoder, and preservation of a
specified stochastic sampling distribution are distinct claims,
each requiring evidence appropriate to its reference process.

\section{Conclusion}
\label{sec:conclusion}

This survey provides a comprehensive review of inference acceleration for diffusion and flow-based generative models, organized around five complementary mechanisms: sampling and trajectory optimization, efficient attention and sparse computation, caching and computation reuse, model compression and quantization, and system optimization and parallel execution. These mechanisms reduce the number of model evaluations, decrease the computational and memory costs of individual evaluations, or improve hardware utilization and execution efficiency. Their growing integration reflects a broader shift from optimizing individual components toward accelerating complete generation pipelines. Across different generation tasks, the output representation determines which computations can be skipped or reused, task dependencies determine how much approximation is tolerable, and the execution path determines the realized speedup. Image generation emphasizes spatial fidelity and semantic alignment, while video generation introduces temporal consistency, long-range dependencies, and real-time constraints. Audio, 3D, world modeling, and multimodal generation introduce further requirements related to temporal continuity, geometric consistency, interactive dynamics, and cross-modal synchronization. In diffusion language models, useful progress also depends on how many tokens can be committed per pass, which intermediate states remain valid, and how much computation is required for refresh and verification. These differences shape the applicability and effectiveness of acceleration mechanisms across tasks.

Looking forward, efficient diffusion inference will increasingly benefit from the joint design of sampling strategies, attention mechanisms, caching policies, numerical precision, and system execution. Evaluations should connect computational savings with generation quality, end-to-end latency, memory consumption, and hardware utilization under clearly specified workloads. Understanding the interactions among acceleration mechanisms and their applicability across generation tasks remains central to developing fast, reliable, and scalable diffusion-based generative systems.

\clearpage
\bibliographystyle{iclr2026_conference}
\bibliography{references}

\end{document}